\documentclass{article}

\usepackage[preprint]{neurips_2023}

\usepackage[utf8]{inputenc} % allow utf-8 input
\usepackage[T1]{fontenc}    % use 8-bit T1 fonts
\usepackage{hyperref}       % hyperlinks
\usepackage{url}            % simple URL typesetting
\usepackage{booktabs}       % professional-quality tables
\usepackage{amsfonts}       % blackboard math symbols
\usepackage{nicefrac}       % compact symbols for 1/2, etc.
\usepackage{microtype}      % microtypography
\usepackage{xcolor}         % colors
\usepackage{amsmath}
\usepackage{wrapfig}
\usepackage{graphicx}
\usepackage[toc,page]{appendix}
\usepackage{float}

\title{An Attention Free Conditional Autoencoder For Anomaly Detection in Cryptocurrencies}

\author{%
  Hugo Inzirillo \\
  Department of Finance and Insurance\\
  CREST-IP Paris\\
  91764 Palaiseau Cedex, France \\
  \texttt{hugo.inzirillo@ip-paris.fr} \\
  \And
    Ludovic De Villelongue \\
  Department of Economics and Finance\\
  Université Paris Dauphine - PSL\\
  75116 Paris, France \\
  \texttt{ludovic.de-villelongue@dauphine.eu} \\
  \
}
\begin{document}

\maketitle

\begin{abstract}
It is difficult to identify anomalies in time series, especially when there is a lot of noise. Denoising techniques can remove the noise but this technique can cause a significant loss of information. To detect anomalies in the time series we have proposed an attention free conditional autoencoder (AF-CA). We started from the autoencoder conditional model on which we added an Attention Free LSTM layer \cite{inzirillo2022attention} in order to make the anomaly detection capacity more reliable and to increase the power of anomaly detection. We compared the results of our Attention Free Conditional Autoencoder with those of an LSTM Autoencoder and clearly improved the explanatory power of the model and therefore the detection of anomaly in noisy time series.
\end{abstract}

\section{Introduction}
%mettre en acant la technique de kelly qui utilise le LSTM qui n'est pas une technique super importante. The objective is to propose an extension of the model proposed by kelly using an Attention Free LSTM layer. We apply this improvement of anomaly dectection.

% Cryptocurrency pricing models can be thought as non linear conditional asset pricing models, where the nonlinearities manifest through a flexible neural network mapping of covariates into betas. 

To identify anomalies in prices on a basket of cryptocurrencies returns, autoencoders can prove themself useful. An autoencoder is a type of neural network in which outputs try to approximate inputs variables. It is divided in two steps, the first one is the encoding part where the input variable is passed inside a small number of neurons in the hidden layers, creating a compressed representation of the input and the second the decoding step which unpack the input representation and maps it to the output layer.  An autoencoder with no other variable than inputs can thus be qualified as a dimension reduction and unsupervised learning device. Self attention mechanisms \citet{attention_free_transformer} embedded within the Transformer allows to increase the efficiency of learning task in machine translation, language understanding and other paradigms.  We introduce an Attention Free LSTM Autoencoder, able to focus on input sequence that will have an impact on prediction. In a latter paper \citet{Autoencoder_asset_pricing_models} introduced non linear functions to identify latent factors of random variables, however they use LSTM layers, which turns out to be less effective than models based on attention mechanisms \citet{Transformer_vs_lstm}. In this paper we look for an architecture to detect anomalies which led us to identify the importance of factors of our time series.
\citet{attention_free_transformer} presented a Attention Free Transformer (AFT) more efficient than Transformer \citet{attention_is_all_u_need} during the learning task, so we started from this point and added and attention free mechanism in the encoder-decoder strucutre to only capture the variables that have a positive impact and minimized the recontruction error. The Attention Free Autoencoder structure we present allows us to better identify anomalies in time series.

\section{Factor Models}
\citet{Autoencoder_asset_pricing_models} proposed a new model to estimates latent factors for stochastic time series. They started from the work of \citet{kerneL_trick_cross_section}  however they introduced non-linear functions to estimate nonlinear conditional exposures and latent factors associated. 

\subsection{Linear Factor Models}
When the autoencoder has one hidden layer and a linear activation function, it is equivalent to the PCA estimator for linear factor models
The static linear factor model is written:
\begin{equation}
    y_t= \beta X_t + u_t,
\end{equation}

 $\left(Y_t\right)_{0{\leq k}{\leq T}}$ is a vector of random variables, $y_t$ is a random variable, $y_t \in \mathbb{R}$  and $X_t$ is a $K$ × $1$ vector of factors, $u_t$ is a $N$ × $1$ vector of idiosyncratic errors (independant of $X_t$), and $\beta$ is the $N$ × $K$ matrix of factor loadings.
The matrix form of the factor model is given by:
\begin{equation}
    Y = \beta X + U.
\end{equation}

\subsection{Non Linear Factor Models}
A gain in precision can be obtained by  incorporating asset-specific covariates in the specification of factor loadings\citet{Autoencoder_asset_pricing_models}. These covariate also indirectly improve estimates of the latent factors themselves. Their model formulation amounts to a combination of two linear models:
\begin{itemize}
\item{A linear specification for the latent factors, that models the factor loadings
as a nonlinear function of covariates, given by:

\begin{equation}
    y_{i,t}=\beta_{i,t-1} X_t + u_{i,t}
\end{equation}}

\item{A linear specification for conditional betas, that models factors as portfolios of individual stock returns, obtained from:

\begin{equation}
    \beta(z_{i,t-1}) = z_{i,t-1}\Gamma
\end{equation}}

\end{itemize}

\section {Autoencoders}
Autoencoders \cite{hinton2006reducing} are a type of artificial neural network trained to reconstruct their input, which forces them to learn a compressed representation of the data. Autoencoders is an unsupervised learning task that can be used for dimensionality reduction as well as feature extraction. There is a close connection between autoencoders and PCA \cite{baldi1989neural} and, by extension, between autoencoders and latent factor asset pricing models \cite{Autoencoder_asset_pricing_models}.

    \subsection{Simple Linear Autoencoder}

 \begin{figure}[!ht]
\centering
\includegraphics[width=0.65\columnwidth]{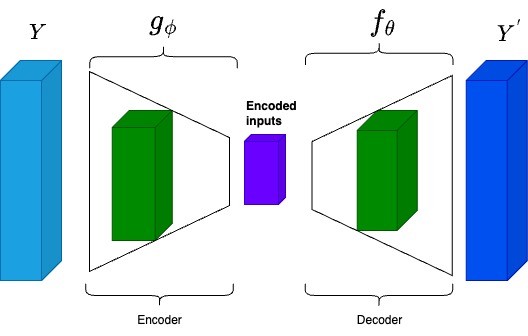}
\caption{Linear Autoencoder }
\label{fig:linear_auto_encoders}
\end{figure}
Let us define $\phi$ and $\theta$ the encoder and decoder set of parameters, respectively. Where  $\phi:=\{\phi^{(0)},\phi^{(1)}\}$ and $\theta:=\{\theta^{(0)},\theta^{(1)}\}$
We can write the one-layer, linear autoencoder with K neurons as:
\begin{equation} 
    \begin{split}
        y_t &=  f_{\theta}(g_{\phi}(y_t)) + u_t, \\
        &=\theta^{(0)}+\theta^{(1)} (\phi^{(0)}+\phi^{(1)}y_t)+u_t,
    \end{split}
\end{equation}
where $\phi^{(1)}$ $\theta^{(1)}$, $\theta^{(0)}$ and $
\phi^{(0)}$ are $K$ × $N$, $N$ ×$K$, $N$ × $1$, and $K$ × $1$ matrices of parameters, respectively. The objective of the model is to minimize the reconstruction error $u_t$ for each time step. $u_t$ is defined as :

\begin{equation}
     u_t = y_t -  f_{\theta}(g_{\phi}(y_t)).
\end{equation}

The parameters of the encoder and the decoder $\phi$ and $\theta$ can be estimated by solving the following optimization problem:

\newcommand\norm[1]{\left\lVert#1\right\rVert}
\begin{equation}
   \min_{\phi, \theta} \sum_{t=1}^{T} \norm{y_t-(\theta^{(0)}+\theta^{(1)} (\phi^{(0)}+\phi^{(1)}y_t))}^{2}.
\end{equation}

    \subsection{Conditional Linear Autoencoders}
     
 \begin{figure}[h!]
\centering
\includegraphics[width=1\columnwidth]{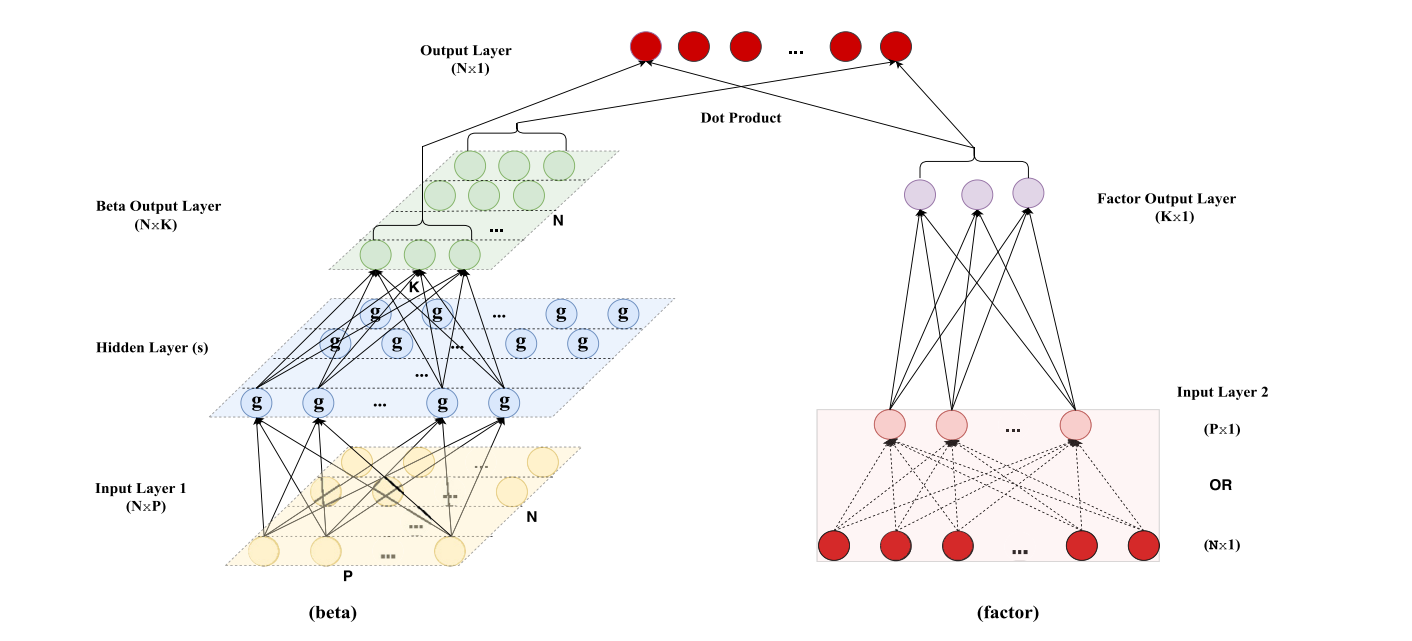}
\caption{Conditional Autoencoders Structure \citet{Autoencoder_asset_pricing_models} }
\label{fig:conditional_autoencoders}
\end{figure}

Figure \ref{fig:conditional_autoencoders}  illustrates the structure of the conditional autoencoder introduced by \citet{Autoencoder_asset_pricing_models}. The left side of the network models factor loadings as a nonlinear function of covariates. This covariates can be characteristics of the time series, such technical indicators. The right side network models factors as portfolios of individual crypto-currency returns.

\begin{equation}
    y_{i,t} =  \beta_{i,t-1}f_t + u_{i,t},
\end{equation}

 The recursive formulation for the nonlinear beta function is: 
\begin{equation}
\begin{split}
    z_{i,t-1}^{(0)} & = z_{i,t-1}, \\
    z_{i,t-1}^{(l)} & = g(b^{(l-1)}+W^{(l-1)} z_{i,t-1}^{(l-1)}),\ l=1,...,L_{\beta}\\
    \beta_{i,t-1} & =b^{(L_{\beta})} + W^{(L_{\beta})} z_{i,t-1}^{(L_{\beta})}.\\
\end{split}
\end{equation}

The recursive
mathematical formulation of the factors is:
\begin{equation}
\begin{split}
    y_t^{(0)} & = y_t \\
    y_t^{(l)} & = \Tilde{g} (\Tilde{b}^{(l-1)} + \Tilde{W}^{(l-1)} y_t^{(l-1)}), \  l=1,...,L_{f} \\
    X_t & =\Tilde{b}^{(L_f)} +\Tilde{W}^{(L_f)} y_t^{(L_f)}
\end{split}
\end{equation}
    
    \subsection{Conditional Attention Free Autoencoders}

\begin{figure}[H]
    \centering
\includegraphics[scale=0.4]{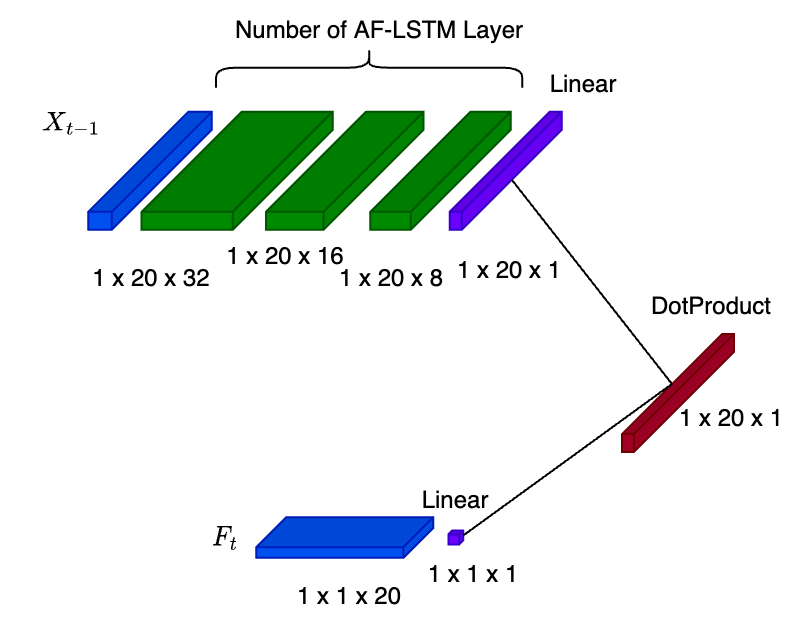}
\caption{Conditional Attention Free Autoencoder}
\label{fig:condition_af_autoencoder}
\end{figure}

    \subsubsection{LSTM}
    
The LSTM framework \citet{lstm} are RNNs with gated mechanisms designed to avoid vanishing gradient. The blocks of LSTM contain 3 non-linear gates; \textit{input gate}, \textit{forget gate} and \textit{output gate}. This architecture embedded a memory for sequences that increase the power of prediction for time series forecasting.  
\begin{itemize}
    \item Input gate ($i_t$): decides which values from the input are used to update the memory,
    \item Forget gate ($f_t$): handles what information to throw away from the block,
    \item Output gate ($o_t$): handles what will be in output based on input and memory gate.
\end{itemize}

During the training step, each iteration provides an update of the model weights proportional to the partial derivative and in some cases the gradient may be vanishingly small and weights may not be updated.
The LSTM networks is defined in Figure \ref{fig:lstm_cell} as:

\begin{figure}[!ht]
    \centering
\includegraphics[scale=0.4]{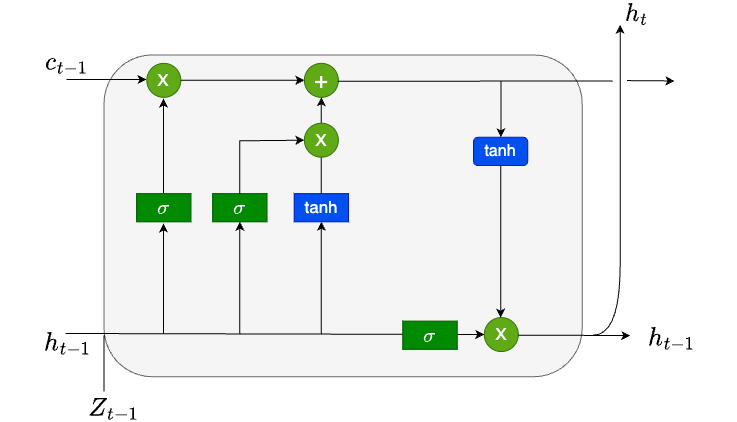}
\caption{LSTM Cell}
\label{fig:lstm_cell}
\end{figure}

\begin{equation}
\label{eq:lstm}
\begin{split}
    f_{t} & = \sigma(W_{f}[h_{t-1},z_{t-1}]+ b_{f})\\
    i_{t} & = \sigma(W_{i}[h_{t-1},z_{t-1}]+ b_{i})\\
    {\tilde c}_{t} & = {\tanh (W_{c}[h_{t-1},z_{t-1}]+ b_{c})}\\
    c_{t} & = f_{t} \odot c_{t-1}+i_{t} \odot {\tilde c}_{t}\\
    o_{t} & = \sigma(W_{o}[h_{t-1},z_{t-1}]+ b_{o})\\ 
    h_{t} & = o_{t} \odot {\tanh(c_{t})} 
\end{split}
\end{equation}
where $ \Theta := \{W_f, W_i, W_c, W_o\}$ is the set of model parameters and  $z_{t-1}$ is the input vector at time t containing all the past information. The number of parameters of each layer in the model should be carefully selected to handle overfitting or underfitting situations.

    \subsubsection{Attention Free Mechanism}
    The transformer architecture has made it possible to develop new models capable of being trained on large dataset while being much better than recurrent neural networks such as LSTM. The Attention Free Transformer \citet{attention_free_transformer} introduces the attention free block, an alternative to attention block \citet{attention_is_all_u_need},  which eliminates the need for dot product self attention. The transformer is an efficient tool to capture the long term dependency. Just like the transformer, the AF Block includes interactions between queries, keys and values. The difference is that the AF block firstly combines key and value together with a set of learned position biases described in \citet{attention_free_transformer}. Then the query is combined with the context vector. Let Q, K, V denote the query, key and value, respectively.

\begin{equation}
    \begin{split}
        Y_t & =  \sigma_q(Q_t) \odot \frac{\sum_{t^{'}=1}^T exp(K_{t^{'}}+w_{t,t^{'}}) \odot V_{t^{'}} }{\sum_{t^{'}=1}^T exp(K_{t^{'}}+w_{t,t^{'}}))}\\
         &= \sigma_q(Q_t) \odot  \sum_{t^{'}=1}^T (\text{Sofmax}(K)\odot V)_{t{'}}
    \end{split}
\end{equation}

where $Q=ZW_q$, $K=ZW_k$ and $V=ZW_v$. The activation function $\sigma_q$  is the sigmoid function and $w_t \in \mathbb{R}^{T\times T}$ is a learned matrix of pair-wise position biases. 

    \subsection{Attention Free LSTM}
    The attention mechanism \citet{attention_is_all_u_need} allows the encoder-decoder models based to be trained only on specific parts of a data sequence that will contribute positively to the prediction of an output, in our case part of the past observations will be kept in order to predict the output of our Conditional Autoencoder. \citet{Autoencoder_asset_pricing_models} 
uses only linear layers to encode and decode the input sequence without necessarily considering temporality and memory. The Attention-Free LSTM Layer \citet{inzirillo2022attention} , brings in addition to the memory effect the possibility to take into account only the necessary input and to filter this input in order to predict the output. The AF-LSTM Layer embeds an attention mechanism as well as an activation function that will allow to eliminate a part of the input sequence that is not useful for the reconstruction of this sequence.

\begin{figure}[!ht]
    \centering
\includegraphics[scale=0.6]{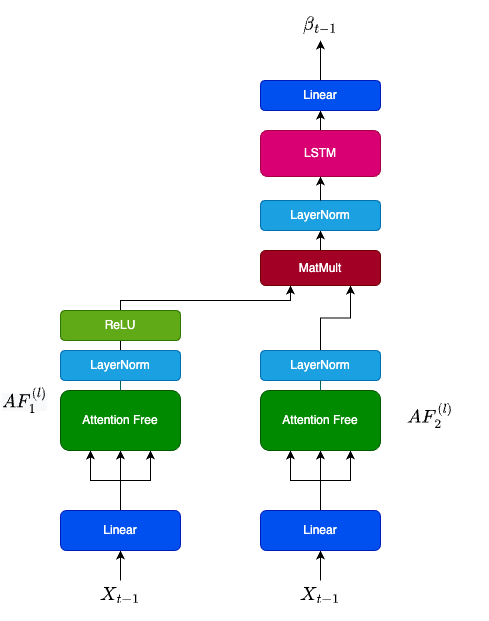}
\caption{Attention-Free LSTM Layer\citet{inzirillo2022attention} }
\label{fig:af_lstm_layer}
\end{figure}

Our $input\_tensor$ for each layer denoted $X$ will be filtered on the left side Figure \ref{fig:af_lstm_layer} using an attention mechanism and an activation function (\textit{ReLU}) to only propagte the revelant information for prediction throught the other hidden layers denoted $l$. Which can be mathematically written such:

\begin{equation}
    \Tilde{X}_{i,t-1}^{(l)} = AF^{(l)}_1(W_{i,1}^{(l)};X_{i,t-1}), \quad  i:=\{1,...,N\}.
\end{equation}

where $X_t \in \mathbb{R}^N$ and $AF^{(l)}_{i,1}$ is the Attention Free of the $l-th$ layer for the $i-th$ time series is defined as:

\begin{equation}
\label{eq:af_mechanism}
\begin{split}
    x_{i,t-1}^{(l)} & = \text{Concat}(W_{x}^{(l)}X_{i,t-1} + b_x^{(l)}),\\
    Q_{i,t-1}^{(l)} & = W_{q}^{(l)}  x_{i,t-1}^{(l)},\\
    K_{i,t-1}^{(l)} &=  W_{k}^{(l)}  x_{i,t-1}^{(l)},\\
    V_{i,t-1}^{(l)} & = W_{v}^{(l)}  x_{i,t-1}^{(l)}, \\
    \eta_{i,t-1}^{(l)} &=  \sum_{t=1}^T \text{Softmax}( K_{i,t-1}^{(l)} )\odot V_{i,t-1}^{(l)},\\
    \Tilde{X}_{i,t-1}^{(l)} &= \sigma( Q_{i,t-1}^{(l)}) \odot \eta_{i,t-1}^{(l)}.\\
\end{split}
\end{equation}

we apply the \textit{Layer Normalization (LN)} function \citet{ba2016layer} was developed to compensate the summed input of reccurent neurons. 

$$
LN(x;\psi;\phi)= \psi \frac{(x-\mu_x)}{\sigma_x}+\phi,
$$

and filter our $\Tilde{X}_{t-1}$ using the \textit{ReLU} activation function, $ReLU(x)=max(0,x)$.

Hence we have:

\begin{equation}
    \Tilde{x}_{t-1} = ReLU(LayerNorm(\Tilde{X}_{t-1})),
\end{equation}

\begin{equation}
\begin{split}
    \Bar{X}_{i,t-1} & = AF_{i,2}^{(l)}(W_{i,2}^{(l)};X_{t-1}),\\
    \Bar{x}_{i,t-1}  & = LN(\Bar{X}_{t-1};\psi^{(l)};\phi^{(l)}).
\end{split}
\end{equation}

The output from the two channels will be multiplied to apply the filter on our input sequence.

\begin{equation}
    \begin{split}
        \eta_{i,t-1}^{(l)} &= ( \Tilde{x}_{i,t-1} \odot \Bar{x}_{i,t-1}),\\
    \zeta_{i,t-1}^{(l)} &= LN( \eta^{(l)};\psi^{(l)};\phi^{(l)}).
    \end{split}
\end{equation}

$\zeta_{t-1}^{(l)}$ will be the input of a parametrized function $g_{w}^{(l)}$ such:

\begin{equation}
    h_{i,t-1}^{(l)} = g_{w}^{(l)}(\zeta_{i,t-1}^{(l)}),
\end{equation}

in Figure \ref{fig:af_lstm_layer} $g_{w}^{(l)}(.)$ is a LSTM Figure \ref{fig:lstm_cell}. Using the output of $g_{w}^{(l)}(.)$ we can estimate the output of the $l-th$ layer $\beta_{i,t-1}^{(l)}$ given by:

\begin{equation}
    \beta_{i,t-1}^{(l_{\beta})} = W_{y}^{(l_{\beta})}h_{i,t-1}^{(l_{\beta})} + b_{y}^{(l_{\beta})}, \quad l:=\{1,...,l_\beta\}.
\end{equation}

\section{Learning task}

\subsection{Datasets}
We take one year of hourly data to perform our analysis on 20 cryptocurrencies. We retrieve the OHLC and volume from \href{https://www.cryptocompare.com/}{CryptoCompare} and calculate five indicators with the  \href{https://pypi.org/project/pandas-ta/}{Pandas TA} : the SMA on 24 hours, the DEMA on 12 hours, the CCI on 24 hours, the AD and the ATR on 24 hours. We then add the time varying hurst exponent by computing log returns, which will be used in the factors part of our models. Finally, we compute the returns on which we will perform training to get predictions. 
In our models, X1 corresponds to the 5 factors values evolution, X2 to the hurst exponent evolution with 1 lag and Y to the returns evolution with 1 lag.
We rank-normalize asset characteristics into the interval (-1,1) for X1 and Y to create tensors scalers.Afterwards, the train set for X1 and Y are fitted and transformed, and the test set for X1 and Y only transformed. The fit method is calculating the mean and variance of each of the features present in the data. The transform method is transforming all the features using the respective mean and variance. The parameters learned by the model using the training data will help to transform the test data.

We finally divide our sample into three time periods to maintain temporal ordering:
\begin{itemize}
    \item The first sample is used for training the model, subject to a specific set of hyperparameter values. It corresponds to 70 percent of 1 year of hourly data.
    \item The second sample is used for validation to tune the hyperparameters.  It corresponds to 15 percent of 1 year of hourly data, beginning after the training set.
    \item The third and last sample is the testing sample used to evaluate a model. It corresponds to the last 15 percent of 1 year of hourly data.
\end{itemize}

\subsection{Autoencoders Types}
For our study we have compared three types of models.
Three autoencoders have been implemented:
\begin{itemize}
    \item Simple autoencoders using linear layers as encoders and decoders. It takes as parameters the input and hidden size or the hidden size and output size. 
    \item LSTM autoencoders using LSTM layers as encoders and decoders. It takes as parameters the input and hidden size or the hidden size and output size. 
    \item AF LSTM autoencoders using AF LSTM layers as encoders and decoders. It takes as parameters the input size, hidden size, the maximum length of the sequence and the output size.
\end{itemize}

\subsection{Autoencoders Dimensions}
The inputs and outputs of the networks are the following:
\begin{itemize}
    \item Input of the Beta Network : 5 which corresponds to the 5 factors in X1.
    \item Input of the Factor Network : 20 which corresponds to the 20 cryptocurrencies selected.
    \item Output of the Beta Network : 1 which is selected as parameter of the models.
    \item Output of the Factor Network : 1 which is selected as parameter of the models.
\end{itemize}

Autoencoders with various degrees of complexity are created:
\begin{itemize}
    \item The simplest, which we denote CA0, uses a single linear layer in both the beta and factor networks.
    \item CA1 which adds a hidden layer with 32 neurons in the beta network.
    \item CA2 which adds a second hidden layer with 16 neurons in the beta network.
    \item CA3 which adds a third hidden layer with 8 neurons in the beta network.
\end{itemize}

\subsection{Regularization}
To prevent overfitting, several regularization techniques are implemented:
\begin{itemize}
    \item Append a penalty to the objective function. In our case the LASSO or ‘l1’ penalization is used. The l1 penalty is set to 0.01 for the simple autoencoders, 0.0016 for the LSTM autoencoders and 0.0000012 for the AF LSTM autoencoders. 
    \item Implement an early stopping tha terminates the optimization when the validation sample errors begin to increase. The tolerance for the early stopping is set to 10. 
    \item Adopt an ensemble approach to train the neural networks. 10 multiple random seeds initialize the neural network estimation and construct model predictions by averaging estimates from all networks. This enhances the stability as different seeds can settle at different optima.
\end{itemize}

\subsection{Optimization}
To reduce the computational intensity of neural networks optimization in computing the loss function, the following algorithms can be added inside the autoencoders models:
\begin{itemize}
    \item A stochastic gradient descent (SGD) with an Adam extension to train a neural network that sacrifices accuracy to accelerate the optimization routine. A critical tuning parameter in SGD is the learning rate, which is set to 0.001 for the three types of autoencoders tested.The number of epochs in our case is set to 200.  
    \item The batch normalization that  performs one activation over one batch on 20 batches.
    \item The layer normalization that normalizes the activations of the previous layer for each given example in a batch independently, rather than across a batch like batch Normalization. It is very effective at stabilizing the hidden state dynamics in recurrent networks. It takes the hidden size of the model as input. 
\end{itemize}

\subsection{Anomaly identification}
We assume that our model is capable of estimating the factors and that its explanatory power is such that the little information it fails to explain through the determined factors indicates the presence of an anomaly or several anomalies. To identify theses anomalies in log returns, we collect the residuals obtained for the train and test sets. The extreme values at a 1 and 5 percent levels are then saved and compared to the log returns observed at the same time. We then consider the log return as abnormal and subject to a market anomaly.

\section{Results}
In this section we will analyze the results for Bitcoin and Ethereum only, which are the two largest cryptocurrencies in terms of market capitalization. The results for the other assets are available in the appendix.

    \subsection{Bitcoin Anomalies Detection}

\begin{figure}[h!]
    \centering
\includegraphics[width=0.85\columnwidth]{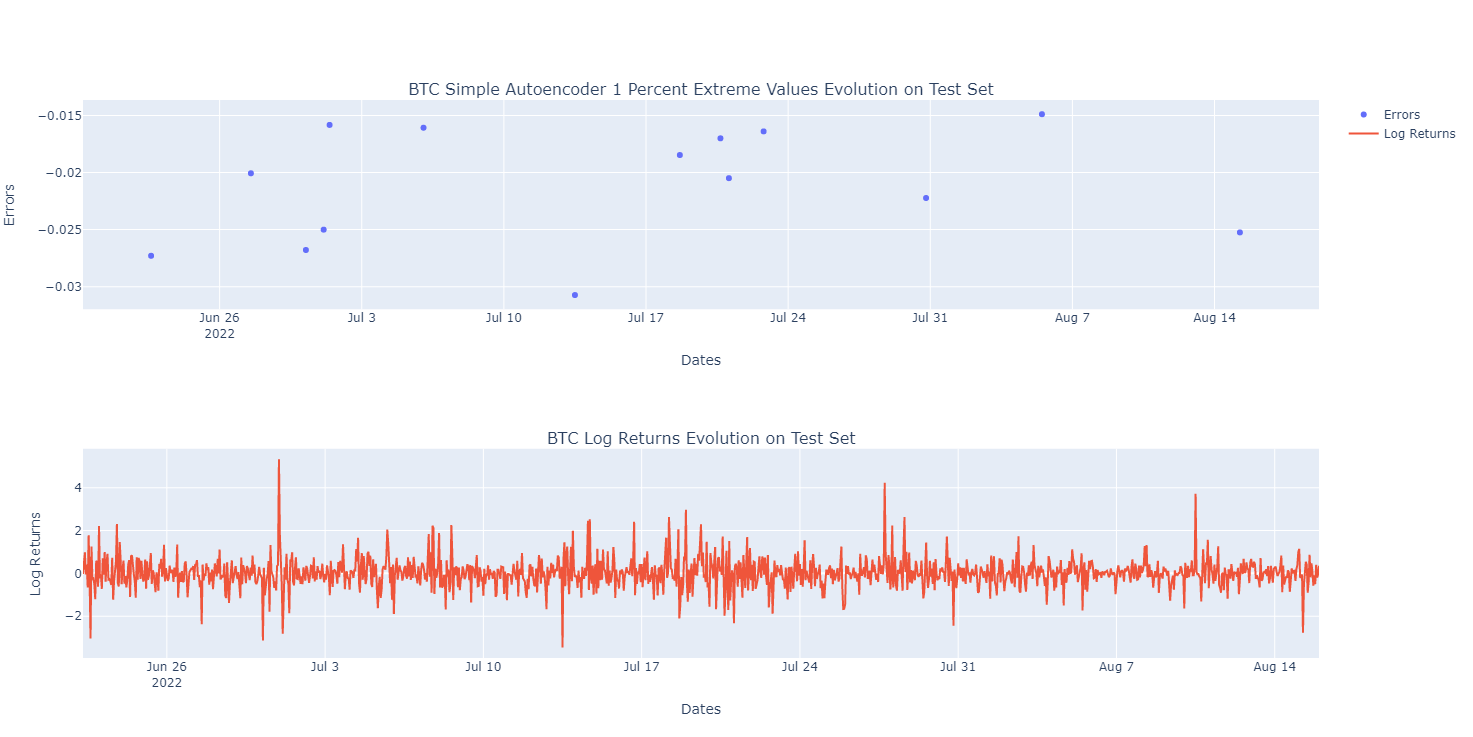}
\caption{BTC Simple Autoencoder 1 Percent Extreme Values Evolution on Test Set}
\label{fig:BTC Simple Autoencoder 1 Percent Extreme Values Evolution on Test Set}
\end{figure}

\begin{figure}[H]
    \centering
\includegraphics[width=0.85\columnwidth]{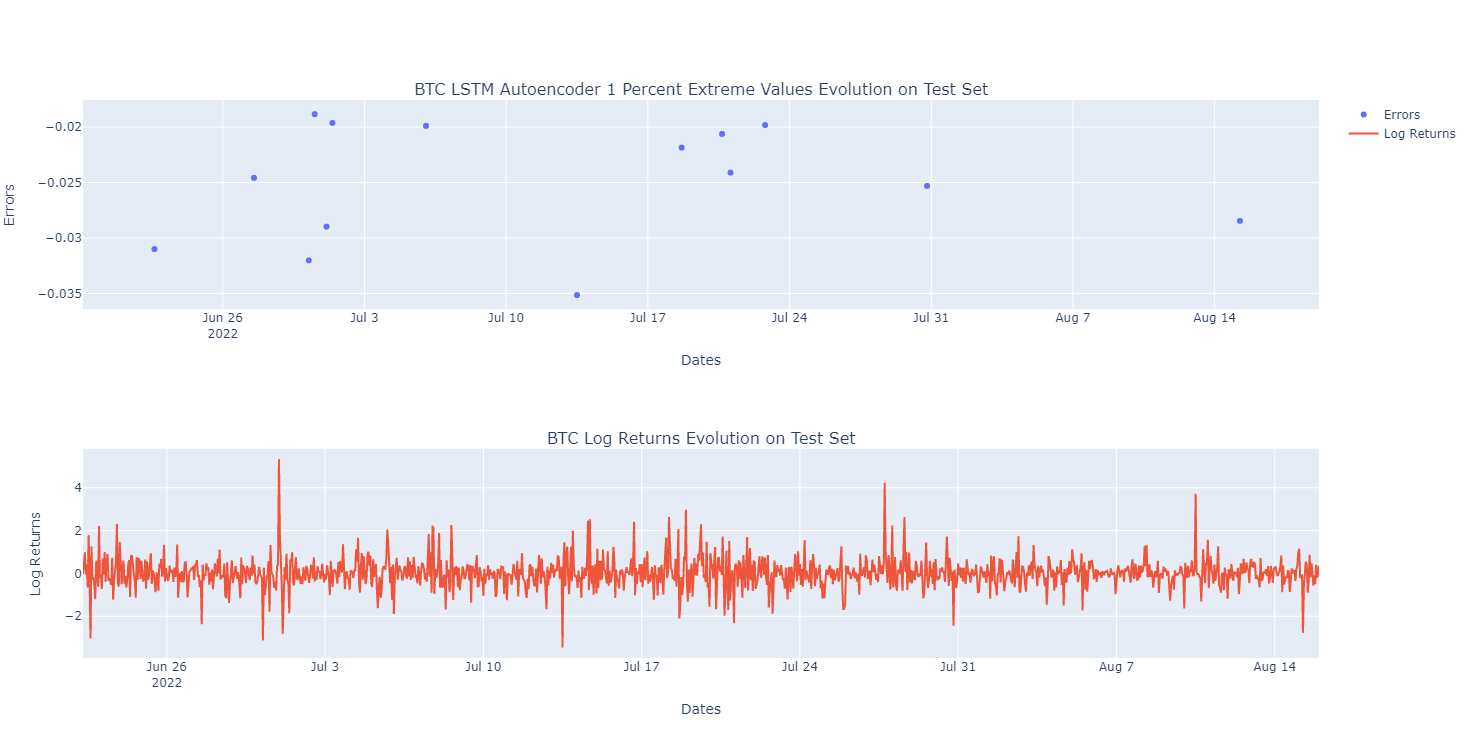}
\caption{BTC LSTM Autoencoder 1 Percent Extreme Values Evolution on Test Set}
\label{fig:BTC LSTM Autoencoder 1 Percent Extreme Values Evolution on Test Set}
\end{figure}

\begin{figure}[H]
\centering
\includegraphics[width=0.85\columnwidth]{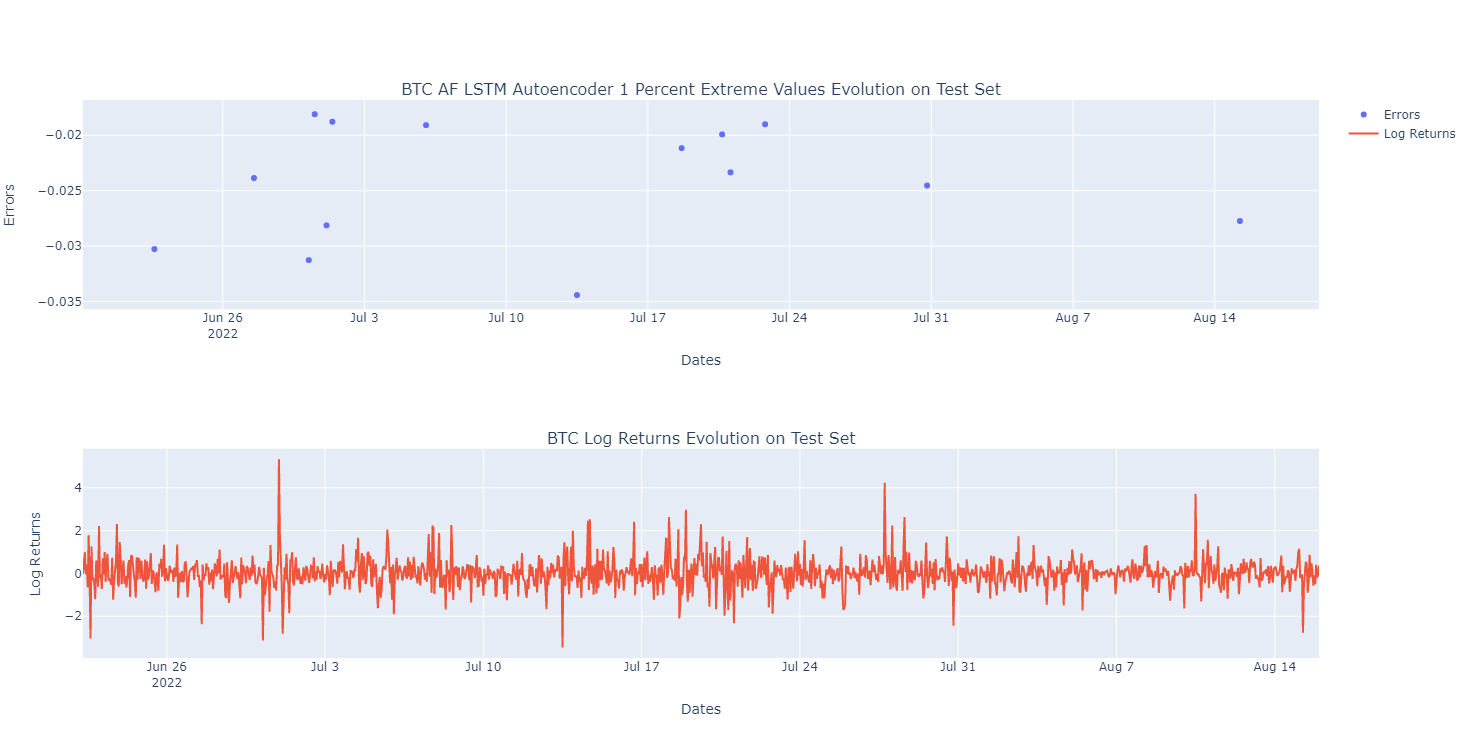}
\caption{BTC AF LSTM Autoencoder 1 Percent Extreme Values Evolution on Test Set}
\label{fig:BTC AF LSTM Autoencoder 1 Percent Extreme Values Evolution on Test Set}
\end{figure}

The Simple Autoencoder model in Fig.\ref{fig:BTC Simple Autoencoder 1 Percent Extreme Values Evolution on Test Set} is able to catch nearly all the plummet phases in log returns. Its squared errors are the smallest of the three models.This model thus focuses more on predictions' quality at the expense of anomalies detection precision. The LSTM autoencoder model in Fig.\ref{fig:BTC LSTM Autoencoder 1 Percent Extreme Values Evolution on Test Set}  is catching an additional anomaly on the 30th of June at 1:00pm, which was ommited by the simple autoencoder model. It is not considering the 5th of August at 12:00pm as an anomaly, which could be seen as an incorrect classification by the simple autoencoder model. Its squared errors are the highest of the three models. This model thus focuses more on anomalies detection precision at the expense of predictions' quality and finally AF LSTM autoencoder model  in Fig.\ref{fig:BTC AF LSTM Autoencoder 1 Percent Extreme Values Evolution on Test Set} benefits from both the power of the Simple Autoencoder and the LSTM Autoencoder in returns prediction and anomaly detection.

\subsection{Ethereum Anomalies Detection}

\begin{figure}[H]
    \centering
\includegraphics[width=0.85\columnwidth]{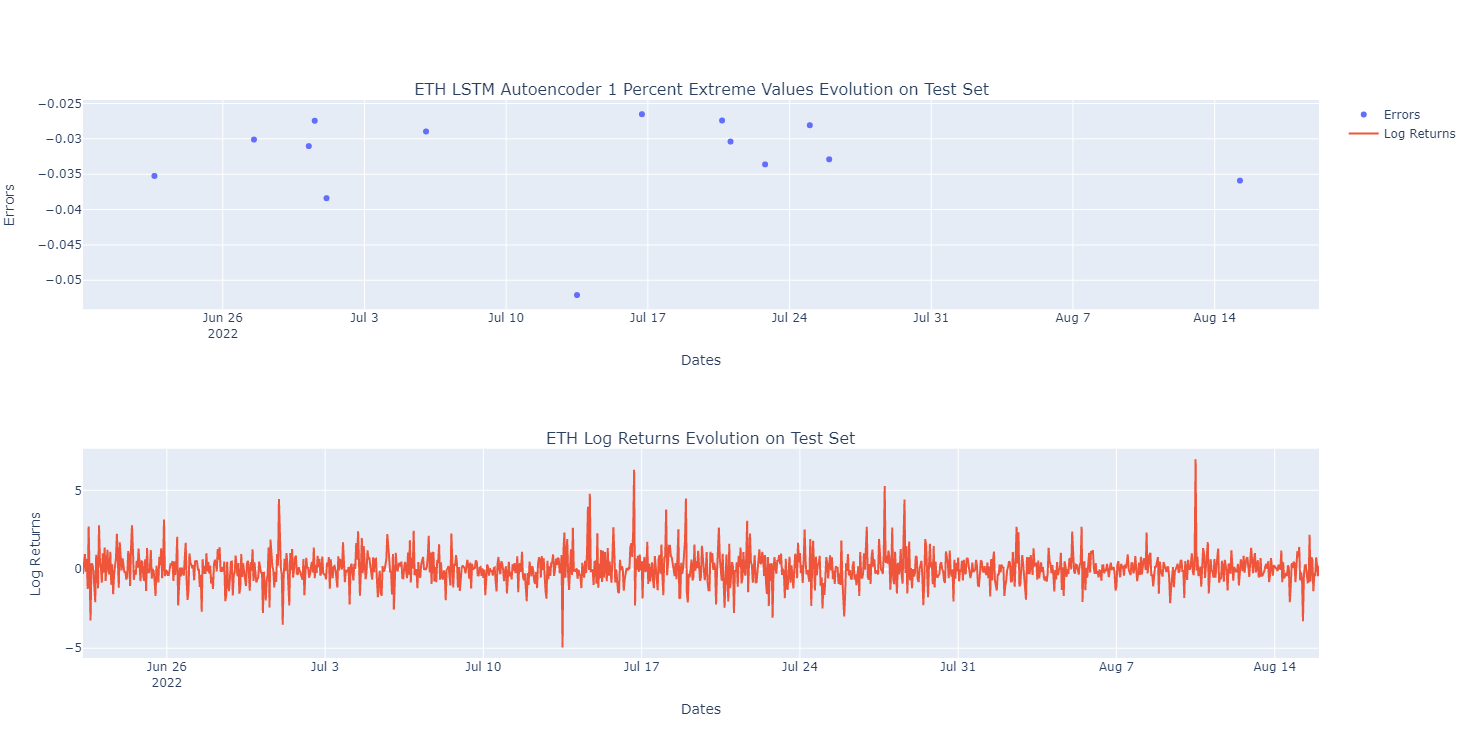}
\caption{ETH Simple Autoencoder 1 Percent Extreme Values Evolution on Test Set}
\label{fig:ETH Simple Autoencoder 1 Percent Extreme Values Evolution on Test Set}
\end{figure}

\begin{figure}[H]
    \centering
\includegraphics[width=0.85\columnwidth]{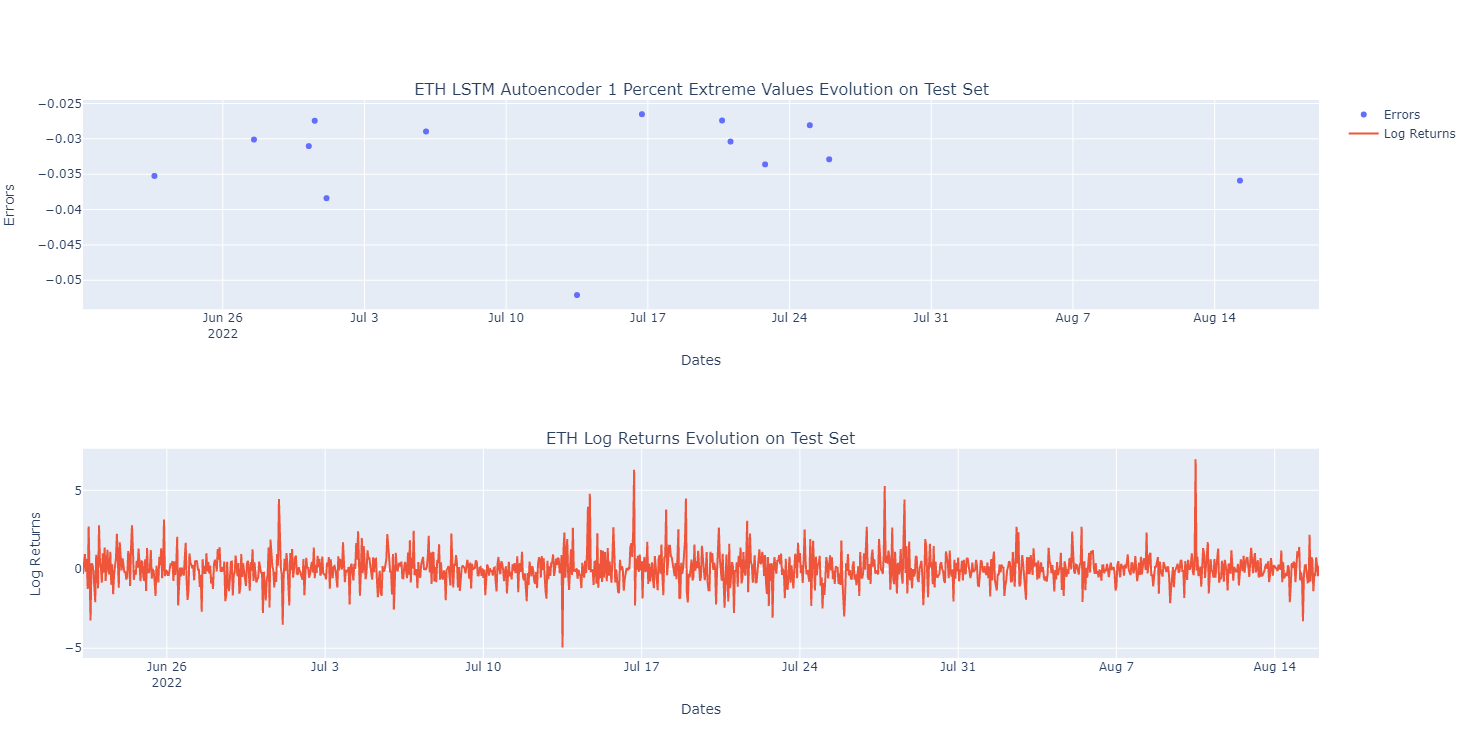}
\caption{ETH LSTM Autoencoder 1 Percent Extreme Values Evolution on Test Set}
\label{fig:ETH LSTM Autoencoder 1 Percent Extreme Values Evolution on Test Set}
\end{figure}

\begin{figure}[H]
    \centering
\includegraphics[width=0.85\columnwidth]{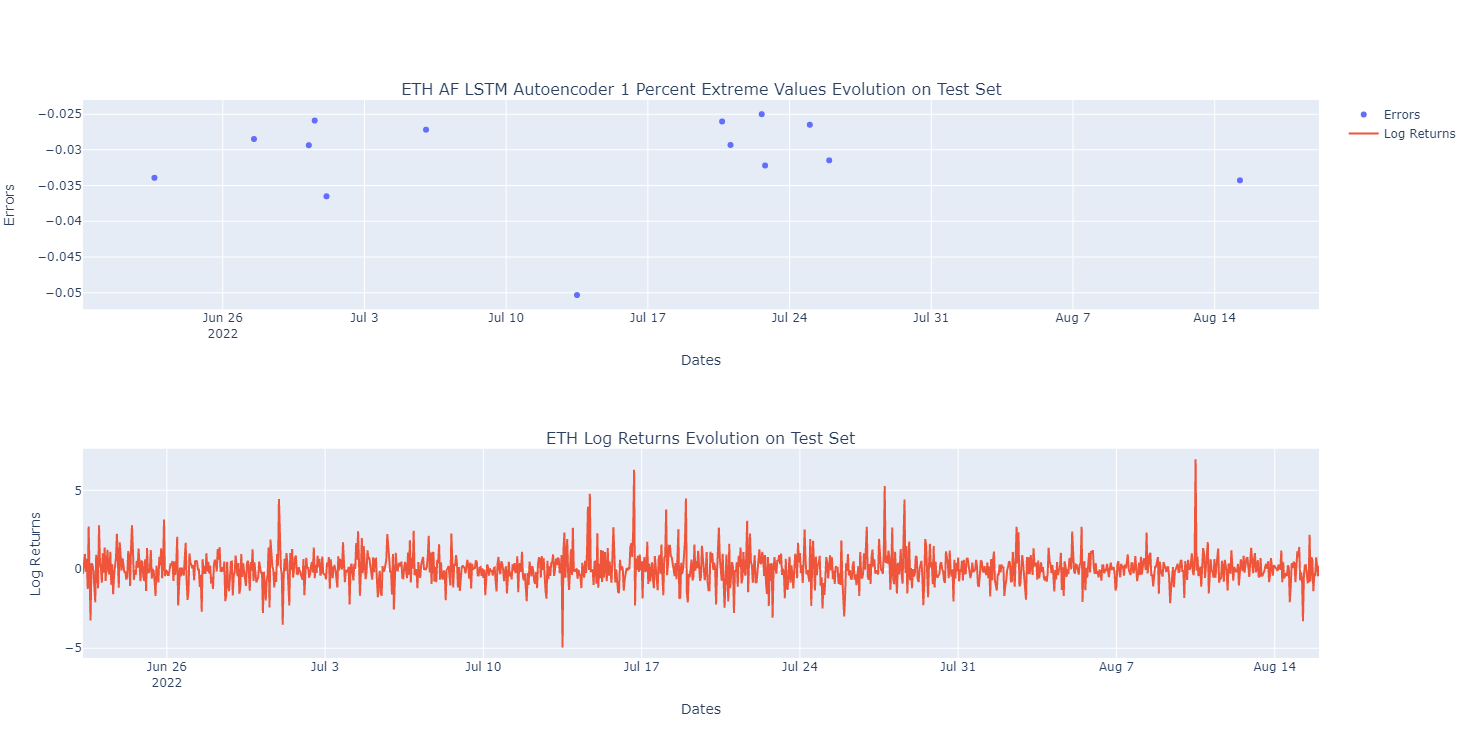}
\caption{ETH AF LSTM Autoencoder 1 Percent Extreme Values Evolution on Test Set}
\label{fig:ETH AF LSTM Autoencoder 1 Percent Extreme Values Evolution on Test Set}
\end{figure}

The Simple Autoencoder model in Fig.\ref{fig:ETH Simple Autoencoder 1 Percent Extreme Values Evolution on Test Set} has a good predictive power. Like the BTC, its squared errors are the smallest of the three models. The LSTM autoencoder model in Fig.\ref{fig:ETH LSTM Autoencoder 1 Percent Extreme Values Evolution on Test Set} is enlighting extra anomalies on the 30th of June at 1:00pm as well as on the 16th of July at 5:00pm. However it does not classify the 22nd of July at 3:00pm as well as the 24th of July at 12:00pm as anomalies. Like the BTC, its squared errors are the highest of the three models and finally AF LSTM autoencoder model  in Fig.\ref{fig:ETH AF LSTM Autoencoder 1 Percent Extreme Values Evolution on Test Set} is not classifying the 16th of July at 5:00pm as an anomaly like the LSTM model but includes the 22nd of July at 3:00pm.

\section{Conclusion}
 Although unsupervised techniques are powerful in detecting outliers, they are subject to overfitting and results that might be unstable.Training multiple models to aggregate the scores can be useful to detect how a change in the network structure can change the algorithms precision in predictions or their capacity to detect anomalies. Outlier detection is a by-product of dimension reduction, explaining the use of autoencoders in our models. The autoencoder techniques can perform non-linear transformations with their non-linear activation function and multiple layers. Instead of providing labels that classify the input features, we compare the prediction of the Autoencoder with the initial input features. It turns out that the AF-CA has a higher explanatory power than the classical autoencoder and thus allows us to better identify anomalies for very noisy time series. In a future work it would be interesting to look at the detection of anomalies in different market regimes, bullish as well as bearish knowing that cryptocurrencies have for main characteristic to be highly volatile assets.

\bibliographystyle{plainnat}
\bibliography{bib}

\begin{thebibliography}{10}
\providecommand{\natexlab}[1]{#1}
\providecommand{\url}[1]{\texttt{#1}}
\expandafter\ifx\csname urlstyle\endcsname\relax
  \providecommand{\doi}[1]{doi: #1}\else
  \providecommand{\doi}{doi: \begingroup \urlstyle{rm}\Url}\fi

\bibitem[Ba et~al.(2016)Ba, Kiros, and Hinton]{ba2016layer}
Jimmy~Lei Ba, Jamie~Ryan Kiros, and Geoffrey~E. Hinton.
\newblock Layer normalization, 2016.

\bibitem[Baldi and Hornik(1989)]{baldi1989neural}
Pierre Baldi and Kurt Hornik.
\newblock Neural networks and principal component analysis: Learning from
  examples without local minima.
\newblock \emph{Neural networks}, 2\penalty0 (1):\penalty0 53--58, 1989.

\bibitem[Gu et~al.(2021)Gu, Kelly, and Xiu]{Autoencoder_asset_pricing_models}
Shihao Gu, Bryan Kelly, and Dacheng Xiu.
\newblock {Autoencoder asset pricing models}.
\newblock \emph{Journal of Econometrics}, 222\penalty0 (1):\penalty0 429--450,
  2021.

\bibitem[Hinton and Salakhutdinov(2006)]{hinton2006reducing}
Geoffrey~E Hinton and Ruslan~R Salakhutdinov.
\newblock Reducing the dimensionality of data with neural networks.
\newblock \emph{science}, 313\penalty0 (5786):\penalty0 504--507, 2006.

\bibitem[Hochreiter and Schmidhuber(1997)]{lstm}
Sepp Hochreiter and Jürgen Schmidhuber.
\newblock Long short-term memory.
\newblock \emph{Neural computation}, 9:\penalty0 1735--80, 12 1997.

\bibitem[Inzirillo and De~Villelongue(2022)]{inzirillo2022attention}
Hugo Inzirillo and Ludovic De~Villelongue.
\newblock An attention free long short-term memory for time series forecasting.
\newblock \emph{arXiv preprint arXiv:2209.09548}, 2022.

\bibitem[Kozak(2019)]{kerneL_trick_cross_section}
Serhiy Kozak.
\newblock Kernel trick for the cross section.
\newblock \emph{SSRN Electronic Journal}, 01 2019.
\newblock \doi{10.2139/ssrn.3307895}.

\bibitem[Vaswani et~al.(2017)Vaswani, Shazeer, Parmar, Uszkoreit, Jones, Gomez,
  Kaiser, and Polosukhin]{attention_is_all_u_need}
Ashish Vaswani, Noam Shazeer, Niki Parmar, Jakob Uszkoreit, Llion Jones,
  Aidan~N. Gomez, Lukasz Kaiser, and Illia Polosukhin.
\newblock Attention is all you need.
\newblock \emph{CoRR}, abs/1706.03762, 2017.

\bibitem[Zeyer et~al.(2019)Zeyer, Bahar, Irie, Schluter, and
  Ney]{Transformer_vs_lstm}
Albert Zeyer, Parnia Bahar, Kazuki Irie, Ralf Schluter, and Hermann Ney.
\newblock A comparison of transformer and lstm encoder decoder models for asr.
\newblock pages 8--15, 12 2019.
\newblock \doi{10.1109/ASRU46091.2019.9004025}.

\bibitem[Zhai et~al.(2021)Zhai, Talbott, Srivastava, Huang, Goh, Zhang, and
  Susskind]{attention_free_transformer}
Shuangfei Zhai, Walter Talbott, Nitish Srivastava, Chen Huang, Hanlin Goh,
  Ruixiang Zhang, and Josh~M. Susskind.
\newblock An attention free transformer.
\newblock \emph{CoRR}, abs/2105.14103, 2021.

\end{thebibliography}

\section{Appendix}

\begin{figure}[H]
    \centering
\includegraphics[width=0.8\columnwidth]{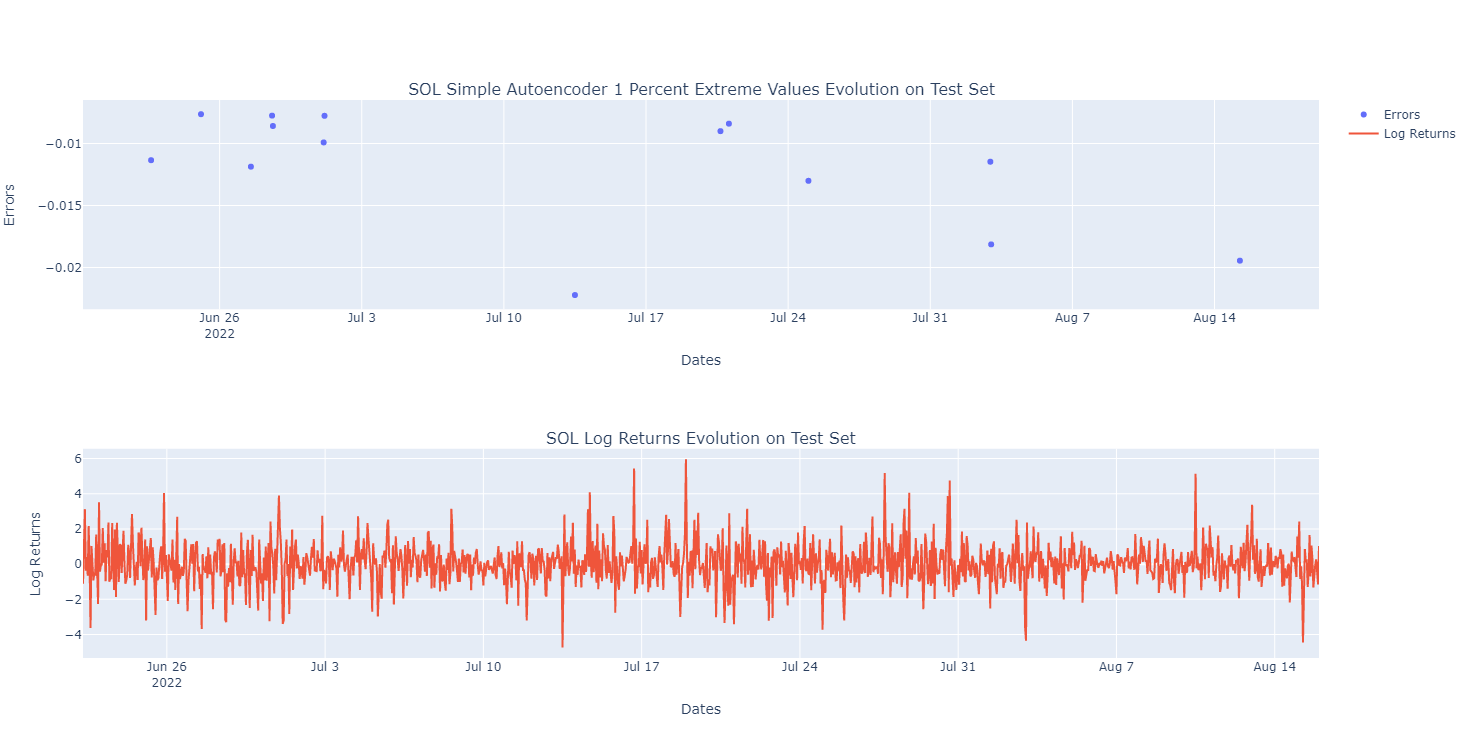}
\caption{XRP Simple Autoencoder 1 Percent Extreme Values Evolution on Test Set}
\label{fig:XRP Simple Autoencoder 1 Percent Extreme Values Evolution on Test Set}
\end{figure}

\begin{figure}[H]
    \centering
\includegraphics[width=0.8\columnwidth]{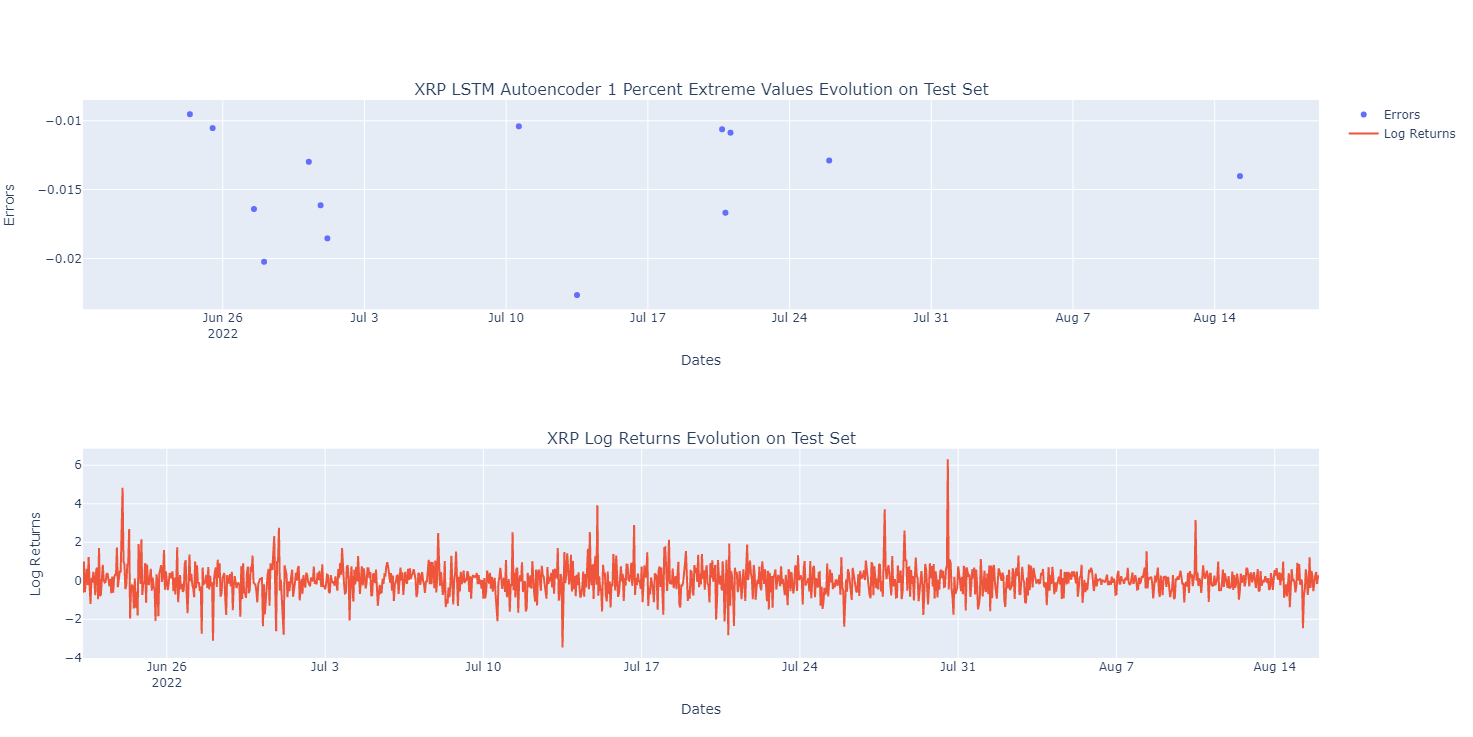}
\caption{XRP LSTM Autoencoder 1 Percent Extreme Values Evolution on Test Set}
\label{fig:XRP LSTM Autoencoder 1 Percent Extreme Values Evolution on Test Set}
\end{figure}

\begin{figure}[H]
\centering
\includegraphics[width=0.8\columnwidth]{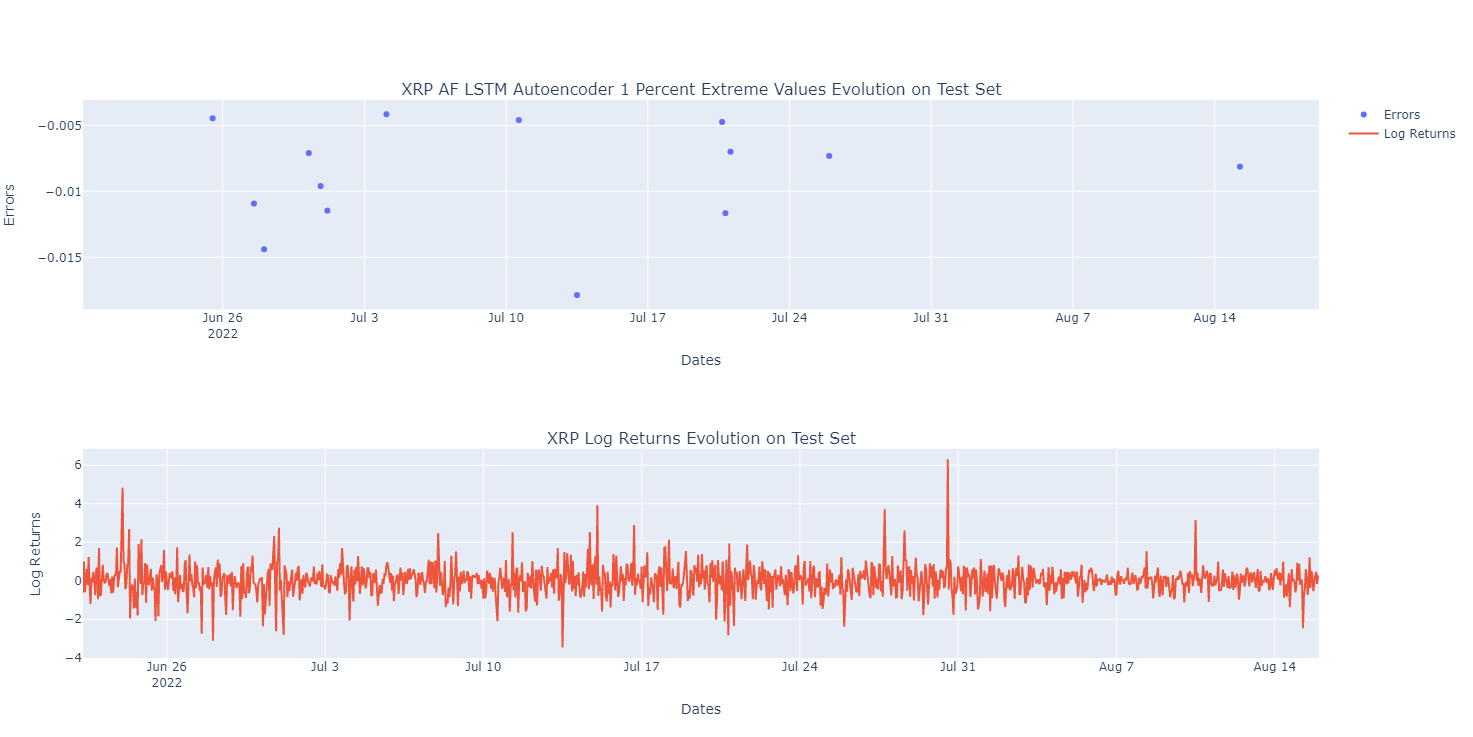}
\caption{XRP AF LSTM Autoencoder 1 Percent Extreme Values Evolution on Test Set}
\label{fig:XRP AF LSTM Autoencoder 1 Percent Extreme Values Evolution on Test Set}
\end{figure}

\begin{figure}[H]
    \centering
\includegraphics[width=0.8\columnwidth]{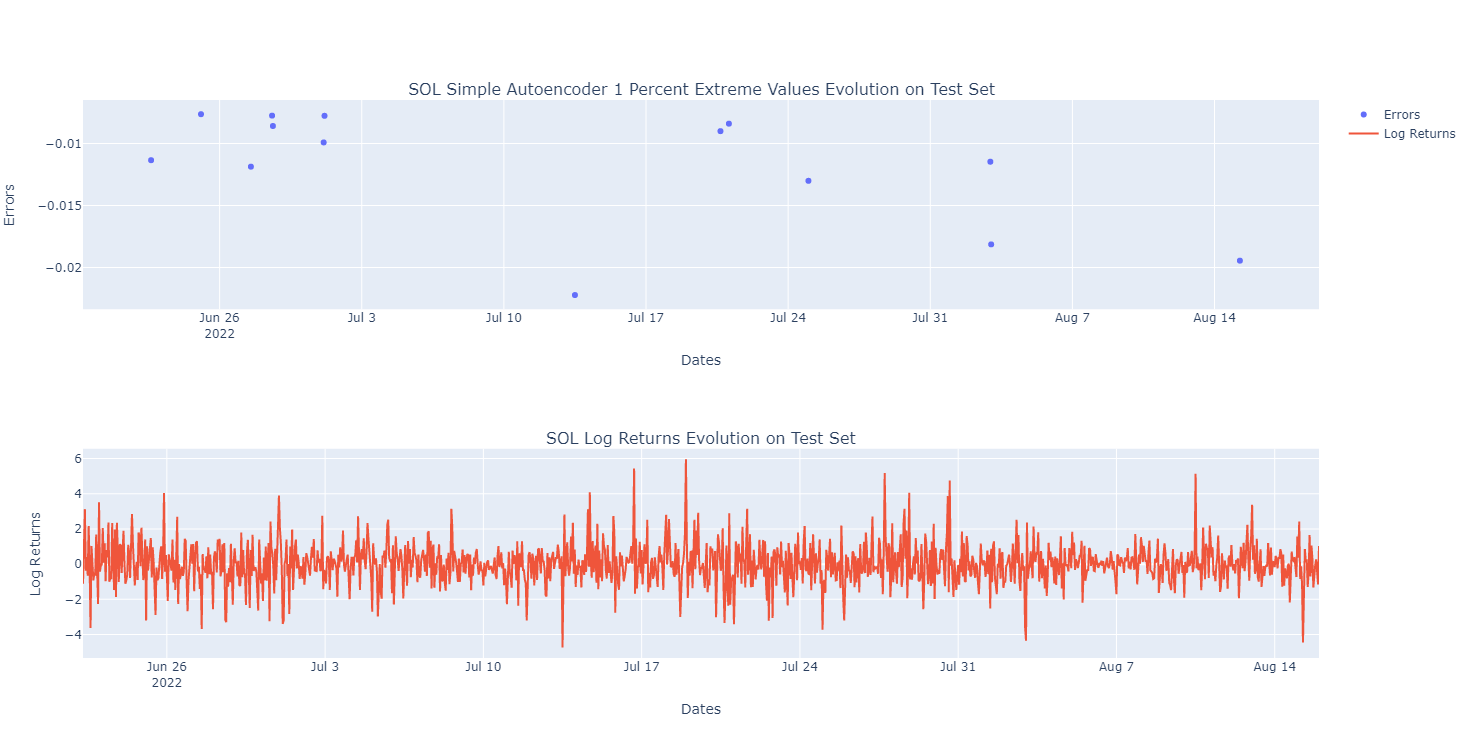}
\caption{SOL Simple Autoencoder 1 Percent Extreme Values Evolution on Test Set}
\label{fig:SOL Simple Autoencoder 1 Percent Extreme Values Evolution on Test Set}
\end{figure}

\begin{figure}[H]
    \centering
\includegraphics[width=0.8\columnwidth]{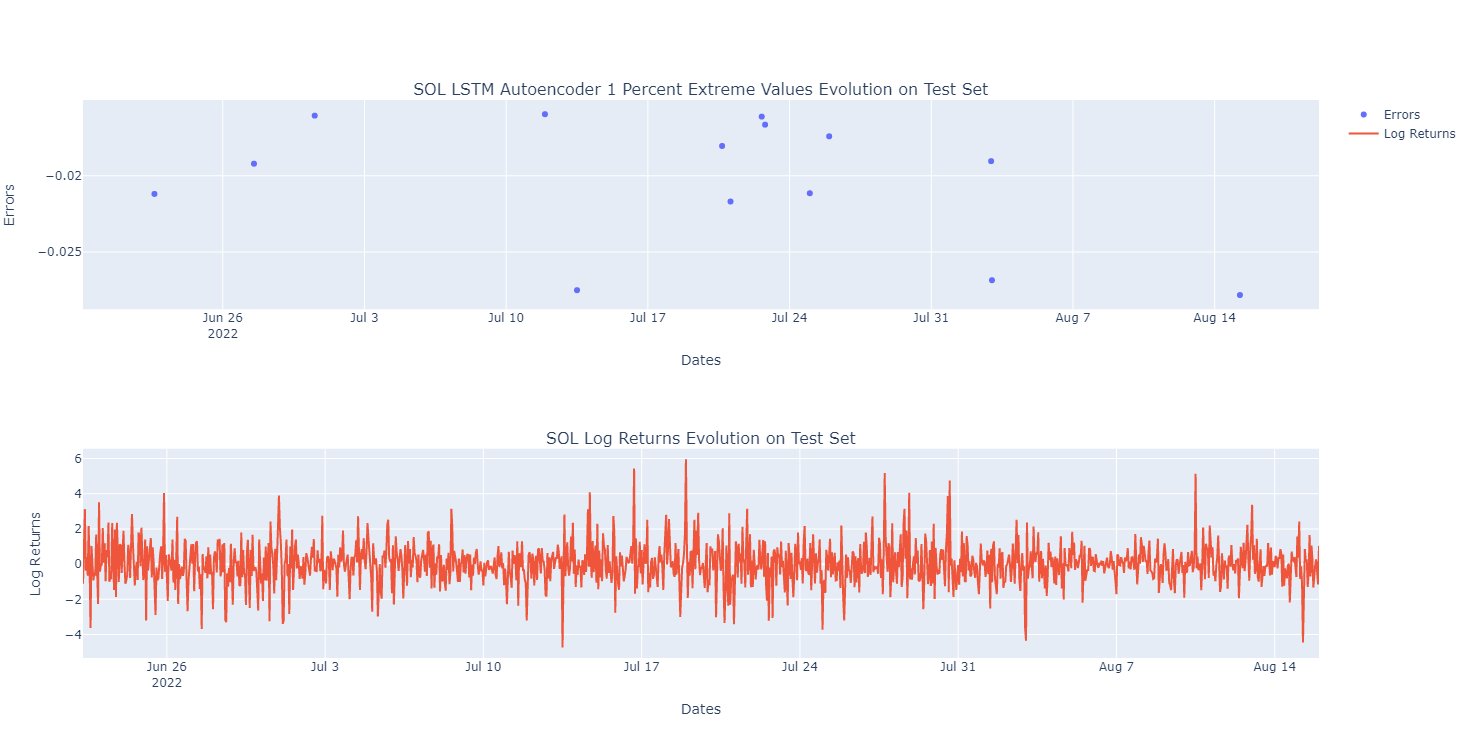}
\caption{SOL LSTM Autoencoder 1 Percent Extreme Values Evolution on Test Set}
\label{fig:SOL LSTM Autoencoder 1 Percent Extreme Values Evolution on Test Set}
\end{figure}

\begin{figure}[H]
    \centering
\includegraphics[width=0.8\columnwidth]{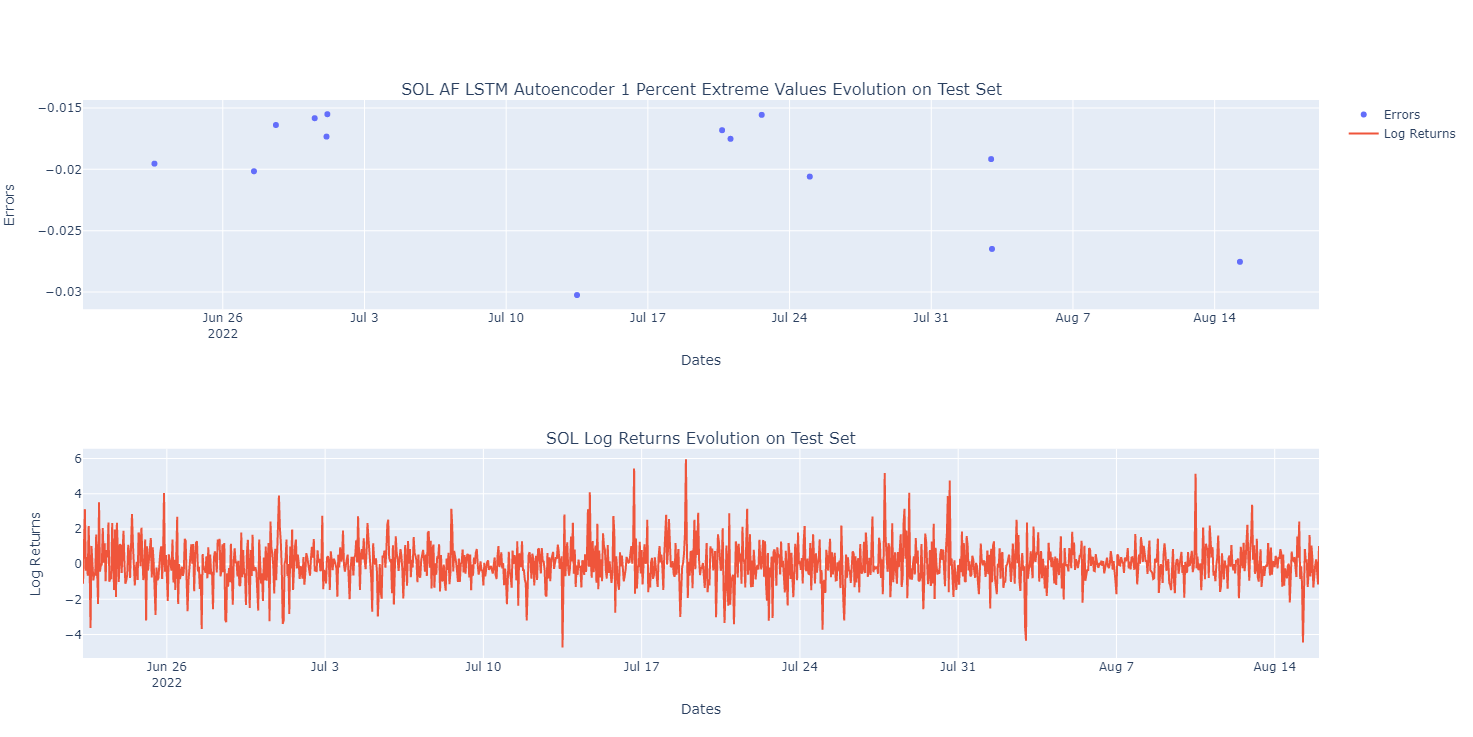}
\caption{SOL AF LSTM Autoencoder 1 Percent Extreme Values Evolution on Test Set}
\label{fig:SOL AF LSTM Autoencoder 1 Percent Extreme Values Evolution on Test Set}
\end{figure}

\begin{figure}[H]
    \centering
\includegraphics[width=0.8\columnwidth]{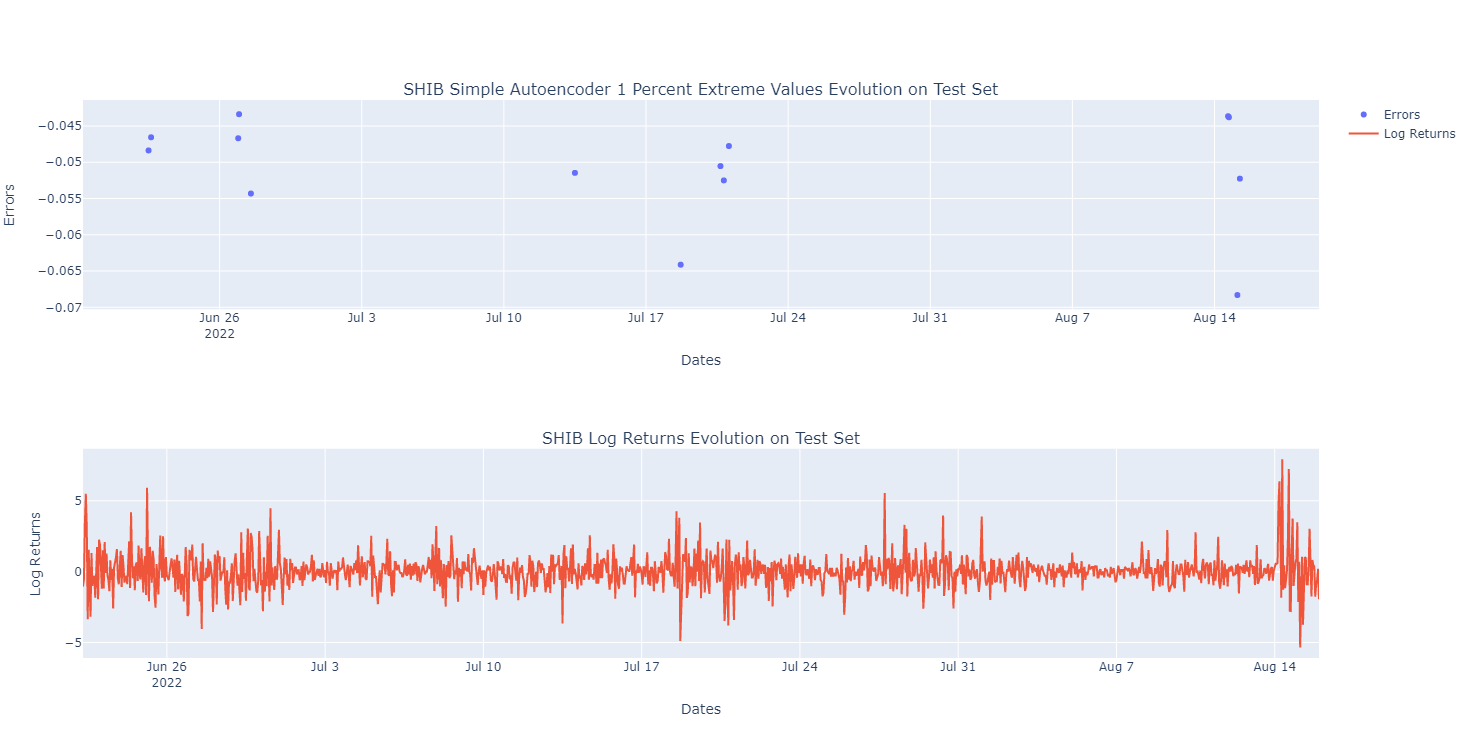}
\caption{SHIB Simple Autoencoder 1 Percent Extreme Values Evolution on Test Set}
\label{fig:SHIB Simple Autoencoder 1 Percent Extreme Values Evolution on Test Set}
\end{figure}

\begin{figure}[H]
    \centering
\includegraphics[width=0.8\columnwidth]{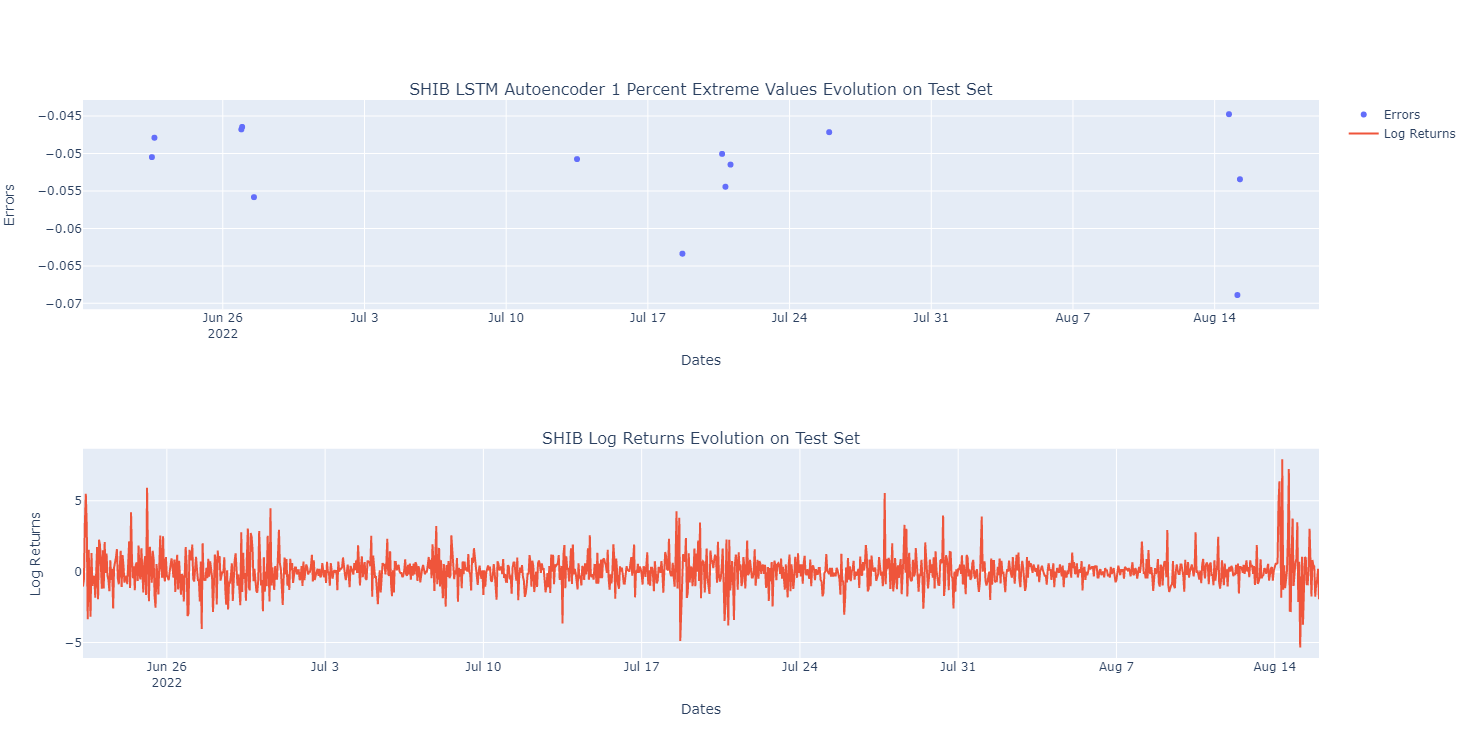}
\caption{SHIB LSTM Autoencoder 1 Percent Extreme Values Evolution on Test Set}
\label{fig:SHIB LSTM Autoencoder 1 Percent Extreme Values Evolution on Test Set}
\end{figure}

\begin{figure}[H]
    \centering
\includegraphics[width=0.8\columnwidth]{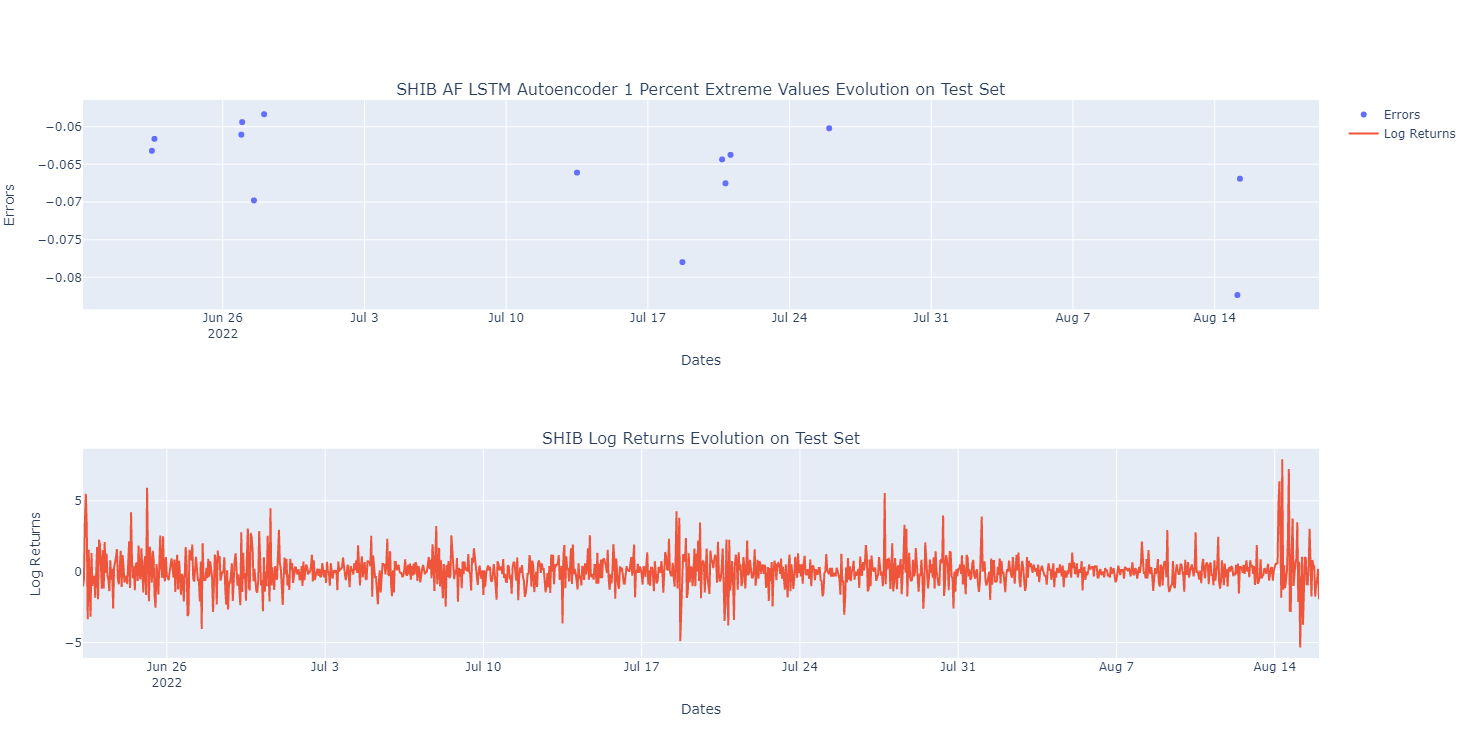}
\caption{SHIB AF LSTM Autoencoder 1 Percent Extreme Values Evolution on Test Set}
\label{fig:SHIB AF LSTM Autoencoder 1 Percent Extreme Values Evolution on Test Set}
\end{figure}

\begin{figure}[H]
    \centering
\includegraphics[width=0.8\columnwidth]{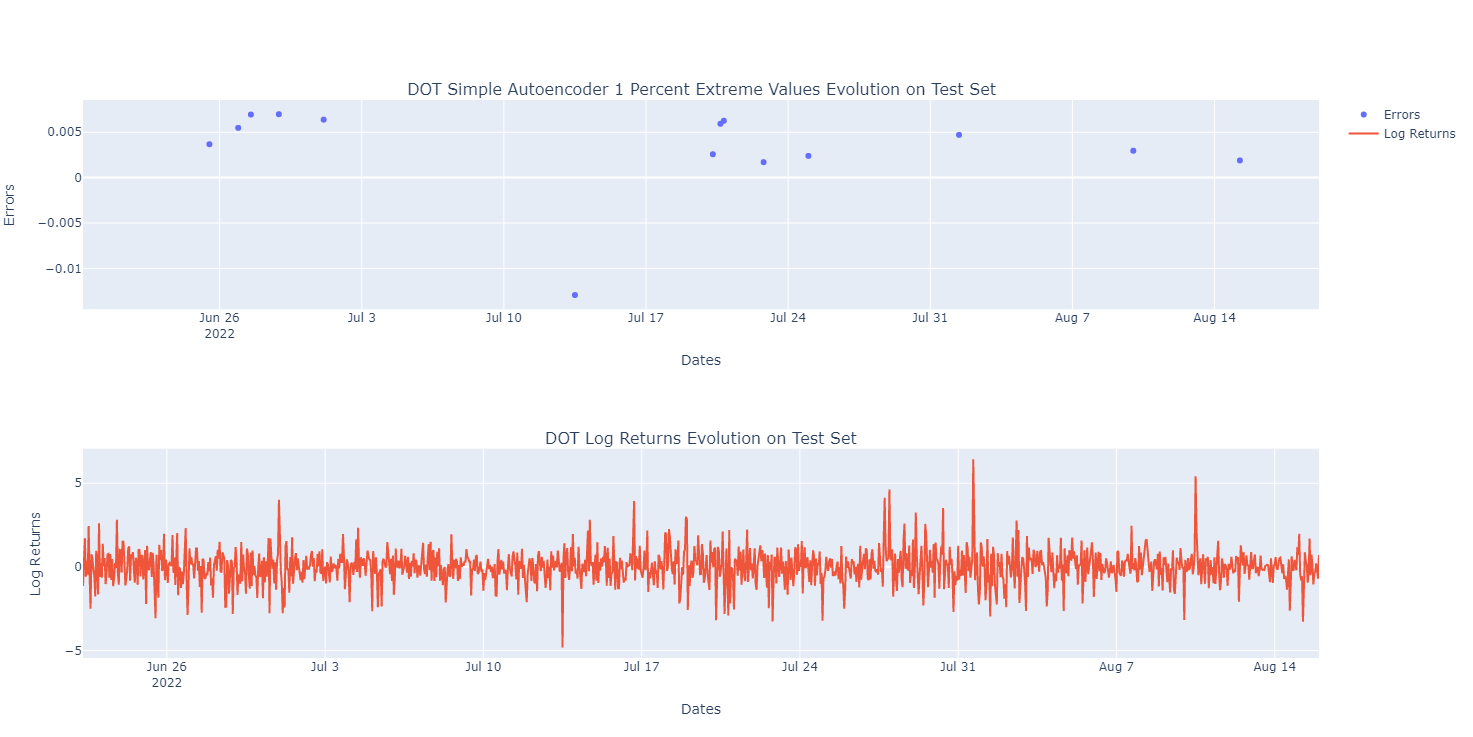}
\caption{DOT Simple Autoencoder 1 Percent Extreme Values Evolution on Test Set}
\label{fig:DOT Simple Autoencoder 1 Percent Extreme Values Evolution on Test Set}
\end{figure}

\begin{figure}[H]
    \centering
\includegraphics[width=0.8\columnwidth]{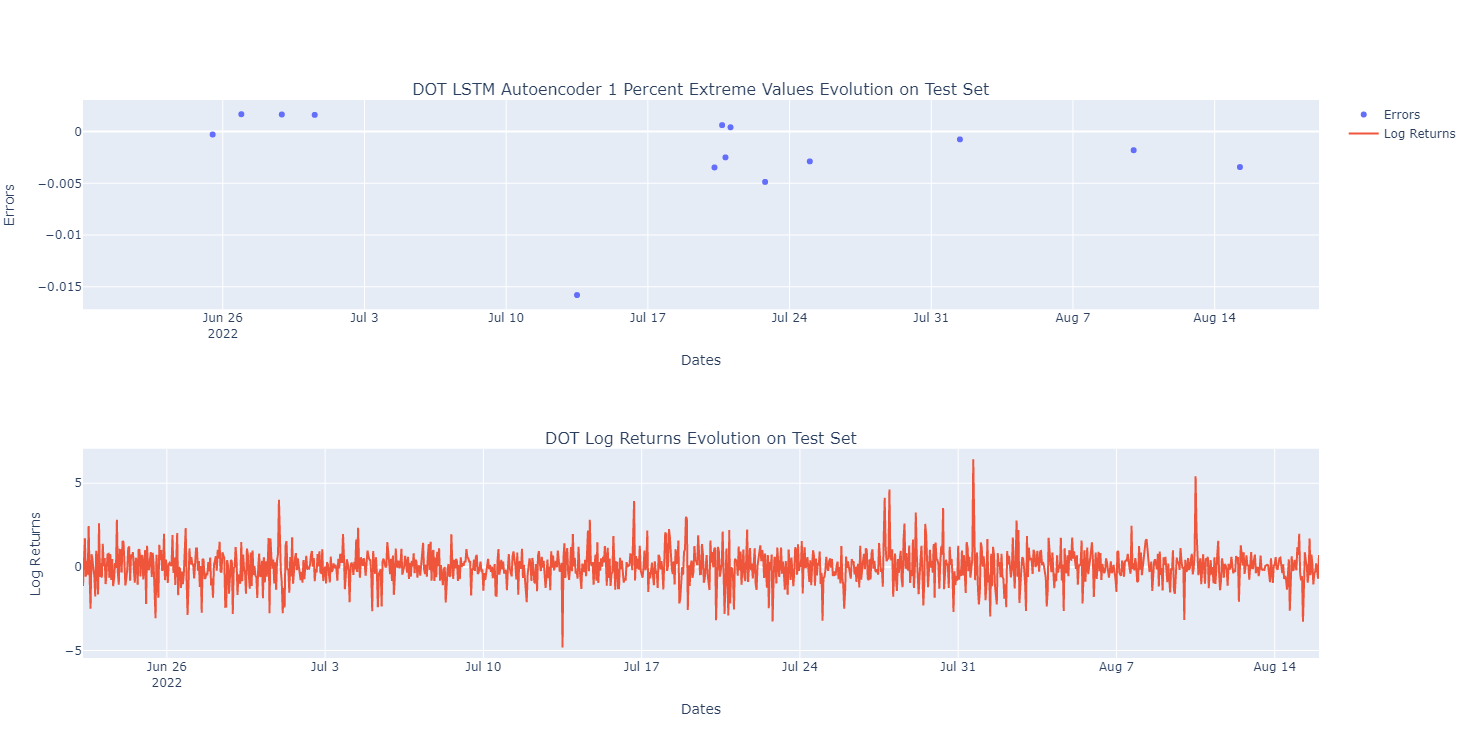}
\caption{DOT LSTM Autoencoder 1 Percent Extreme Values Evolution on Test Set}
\label{fig:DOT LSTM Autoencoder 1 Percent Extreme Values Evolution on Test Set}
\end{figure}

\begin{figure}[H]
    \centering
\includegraphics[width=0.8\columnwidth]{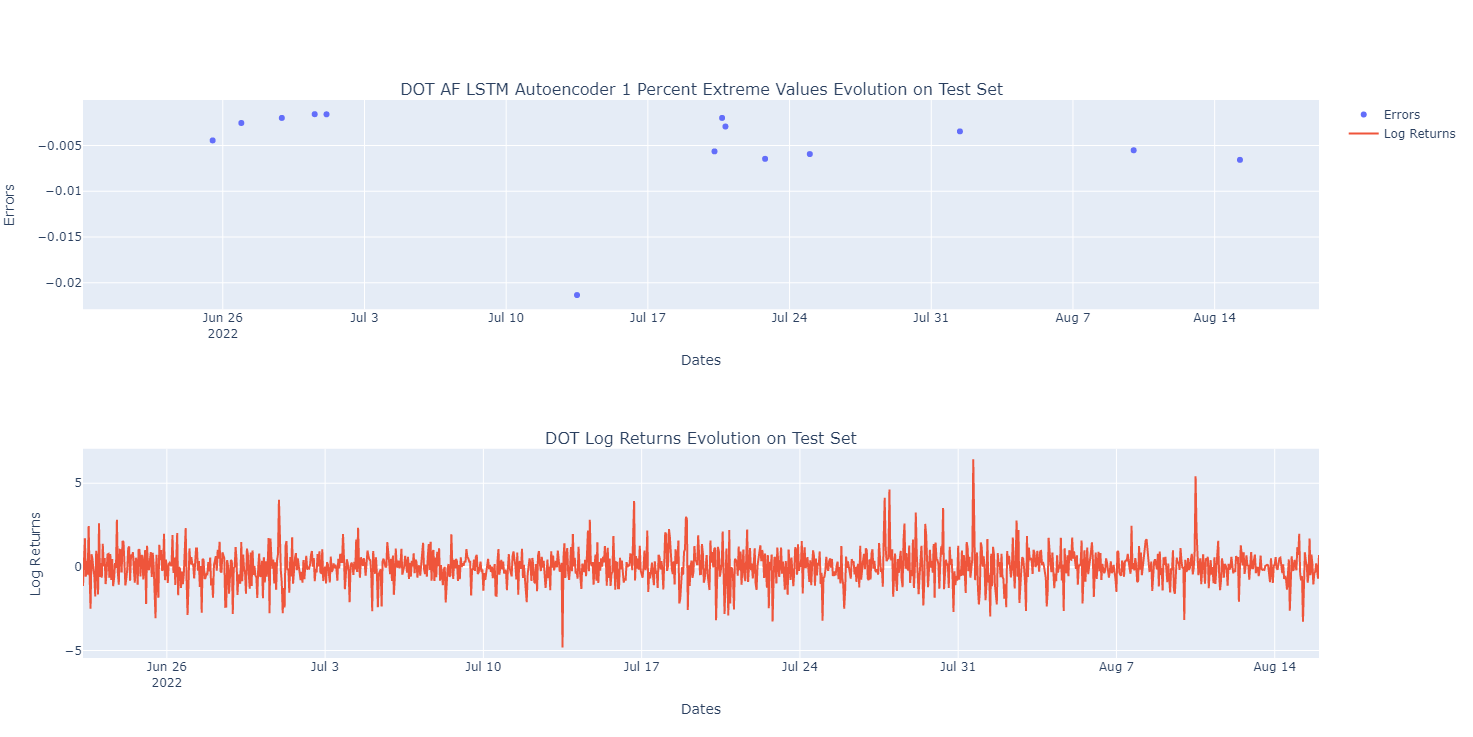}
\caption{DOT AF LSTM Autoencoder 1 Percent Extreme Values Evolution on Test Set}
\label{fig:DOT AF LSTM Autoencoder 1 Percent Extreme Values Evolution on Test Set}
\end{figure}

\begin{figure}[H]
    \centering
\includegraphics[width=0.8\columnwidth]{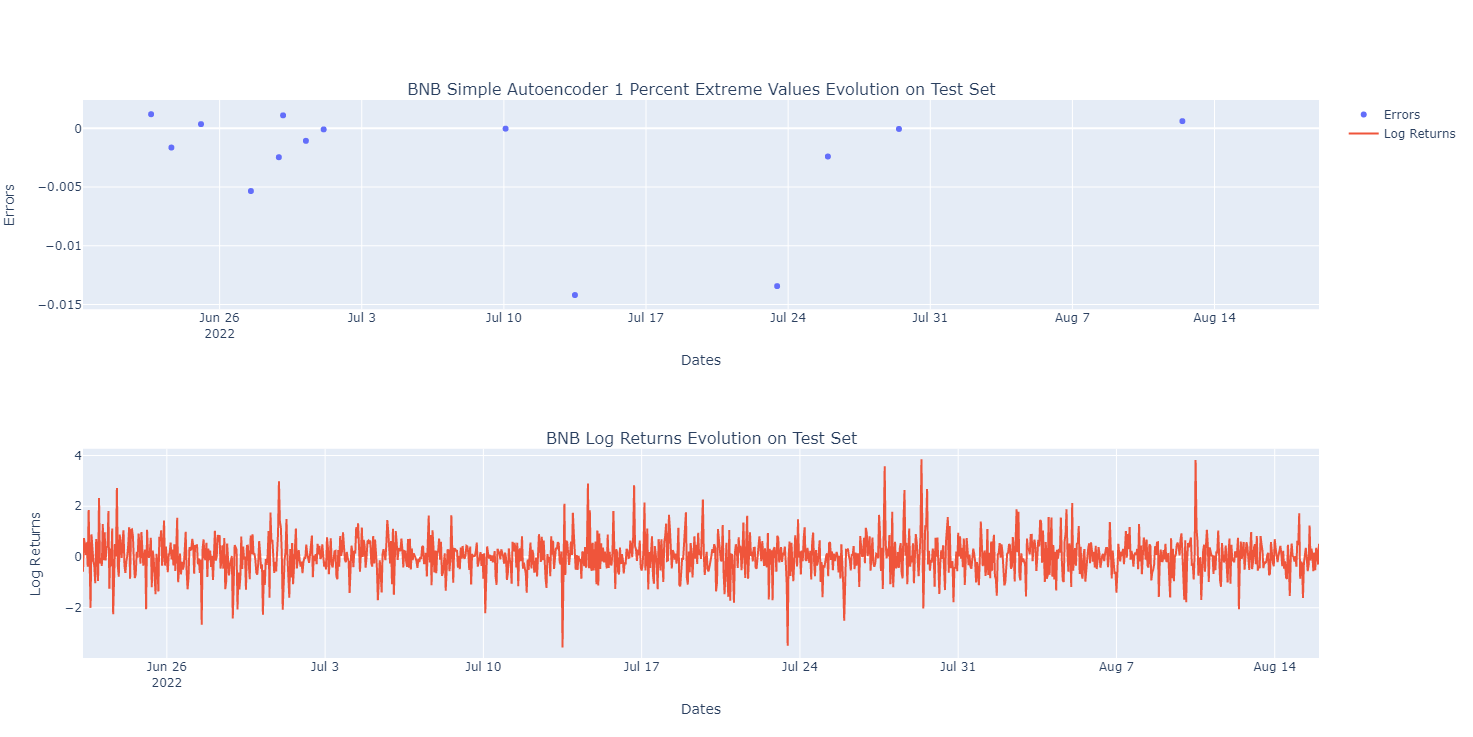}
\caption{BNB Simple Autoencoder 1 Percent Extreme Values Evolution on Test Set}
\label{fig:BNB Simple Autoencoder 1 Percent Extreme Values Evolution on Test Set}
\end{figure}

\begin{figure}[H]
    \centering
\includegraphics[width=0.8\columnwidth]{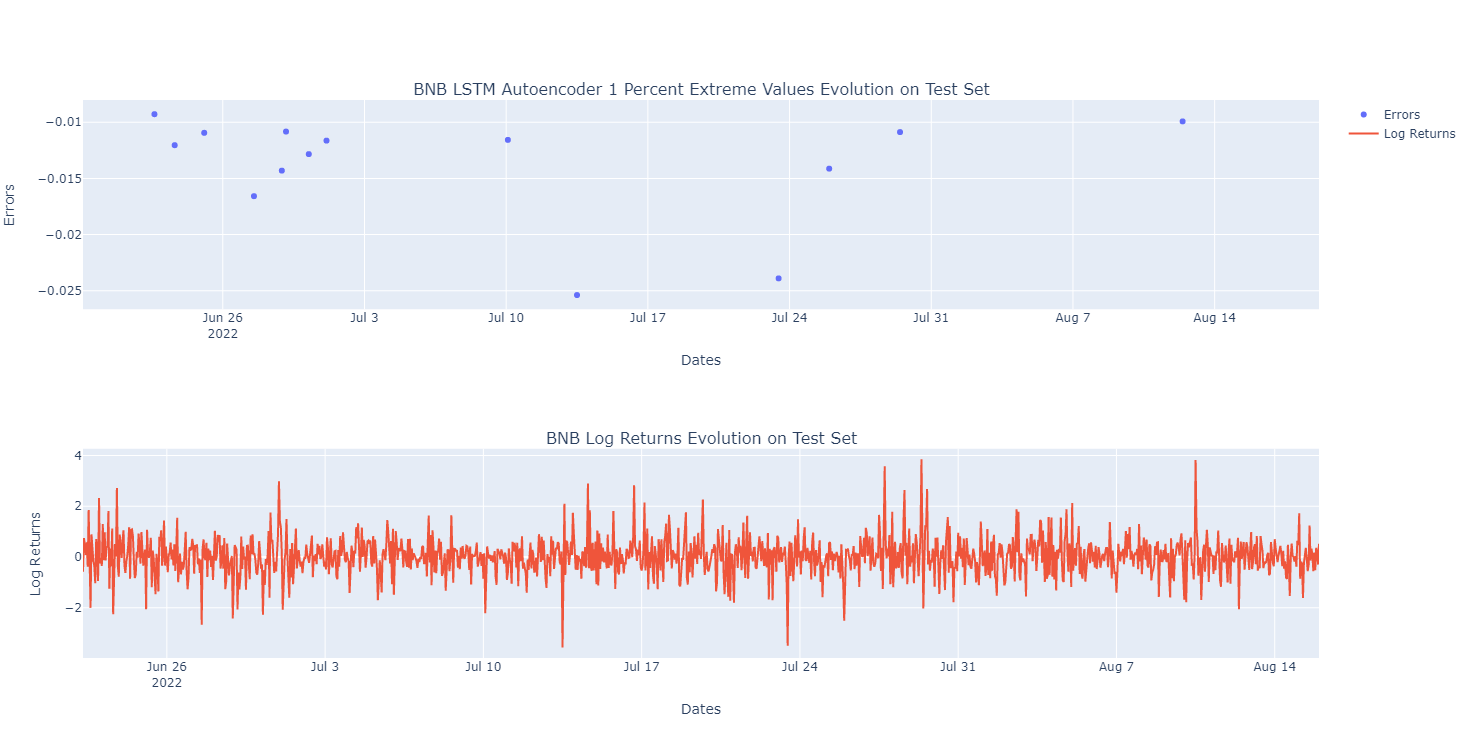}
\caption{BNB LSTM Autoencoder 1 Percent Extreme Values Evolution on Test Set}
\label{fig:BNB LSTM Autoencoder 1 Percent Extreme Values Evolution on Test Set}
\end{figure}

\begin{figure}[H]
    \centering
\includegraphics[width=0.8\columnwidth]{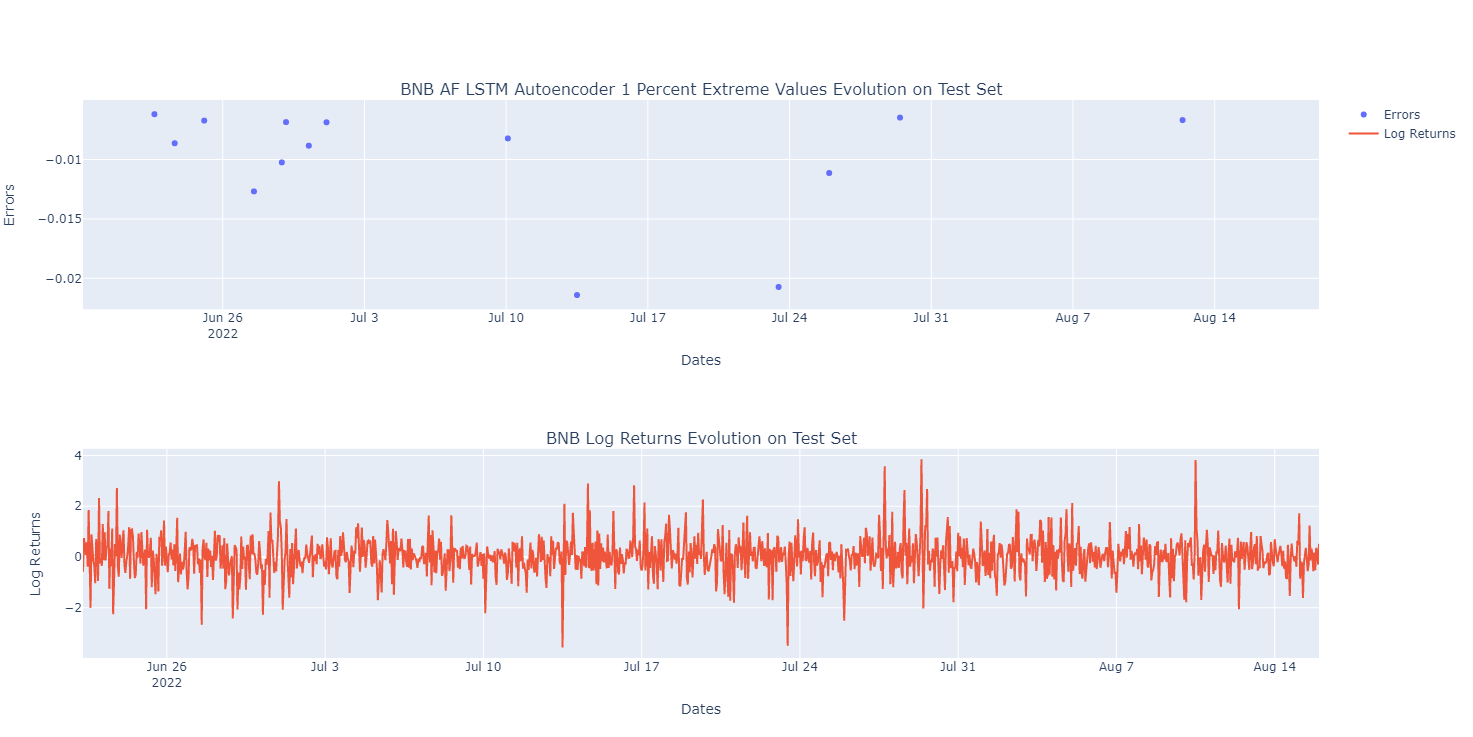}
\caption{BNB AF LSTM Autoencoder 1 Percent Extreme Values Evolution on Test Set}
\label{fig:BNB AF LSTM Autoencoder 1 Percent Extreme Values Evolution on Test Set}
\end{figure}

\begin{figure}[H]
    \centering
\includegraphics[width=0.8\columnwidth]{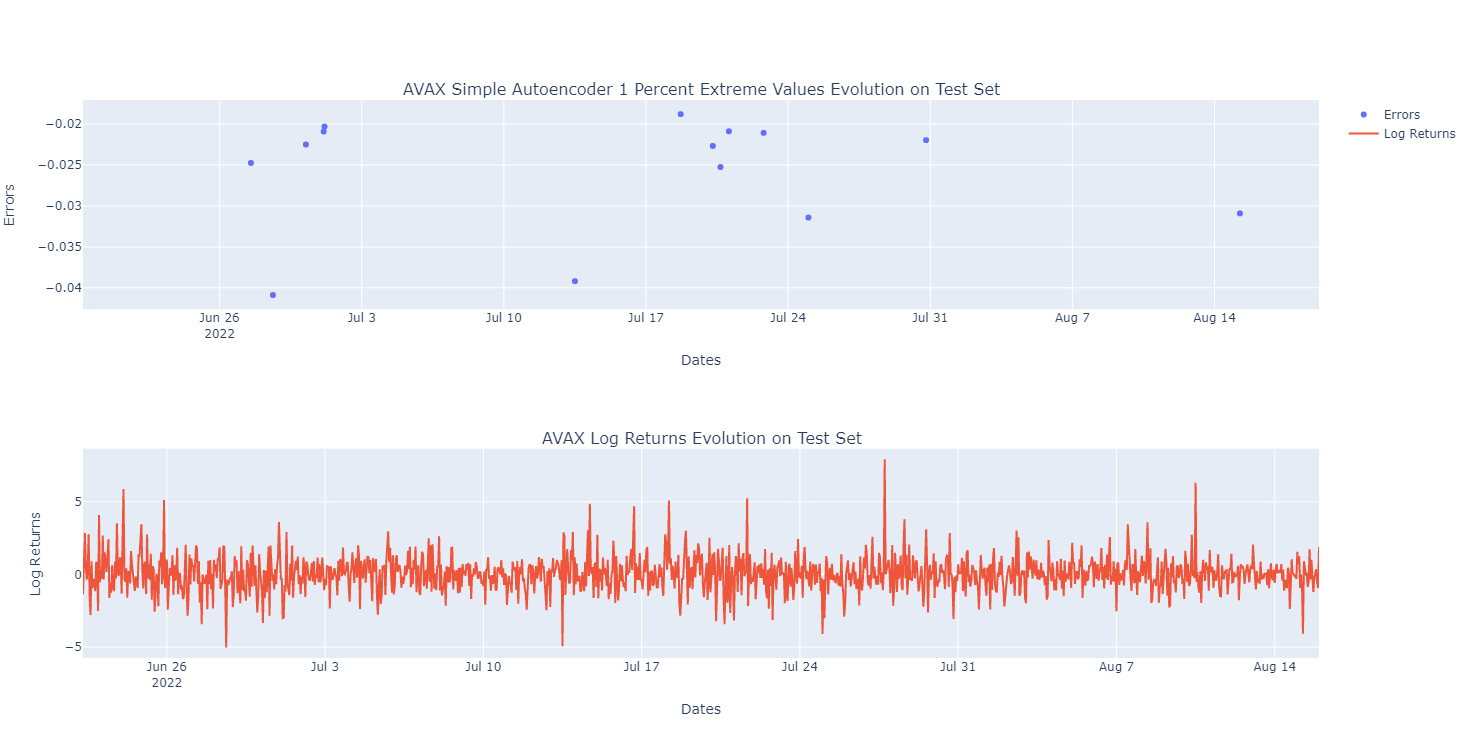}
\caption{AVAX Simple Autoencoder 1 Percent Extreme Values Evolution on Test Set}
\label{fig:AVAX Simple Autoencoder 1 Percent Extreme Values Evolution on Test Set}
\end{figure}

\begin{figure}[H]
    \centering
\includegraphics[width=0.8\columnwidth]{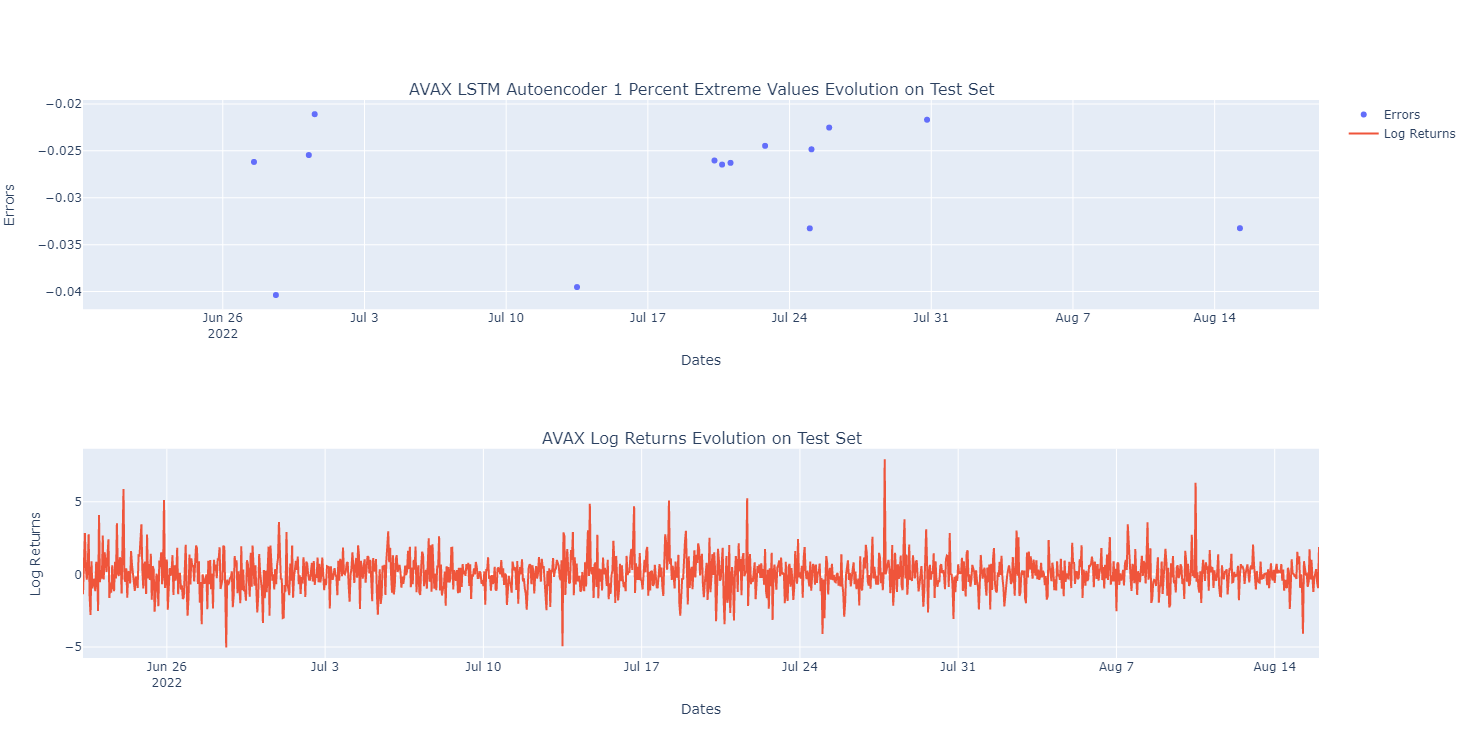}
\caption{AVAX LSTM Autoencoder 1 Percent Extreme Values Evolution on Test Set}
\label{fig:AVAX LSTM Autoencoder 1 Percent Extreme Values Evolution on Test Set}
\end{figure}

\begin{figure}[H]
    \centering
\includegraphics[width=0.8\columnwidth]{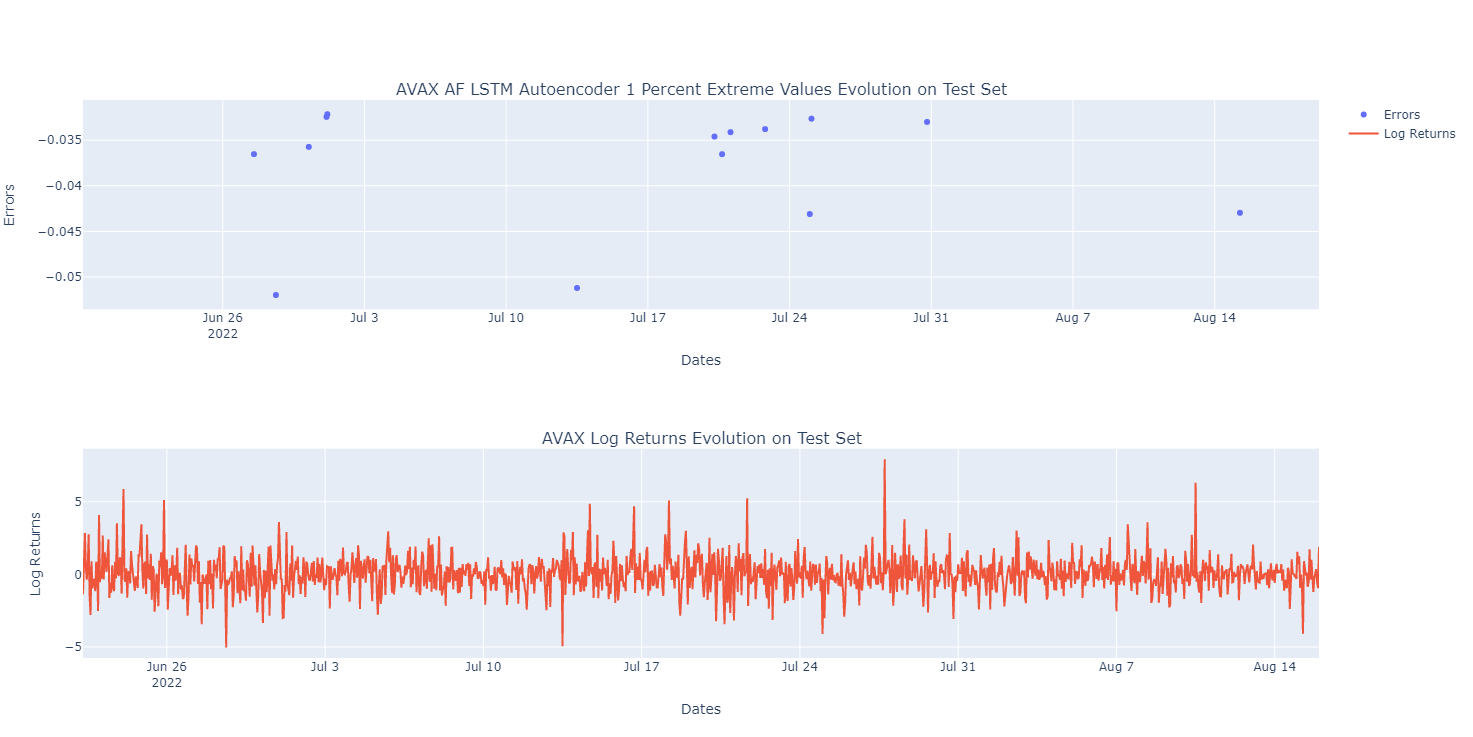}
\caption{AVAX AF LSTM Autoencoder 1 Percent Extreme Values Evolution on Test Set}
\label{fig:AVAX AF LSTM Autoencoder 1 Percent Extreme Values Evolution on Test Set}
\end{figure}

\begin{figure}[H]
    \centering
\includegraphics[width=0.8\columnwidth]{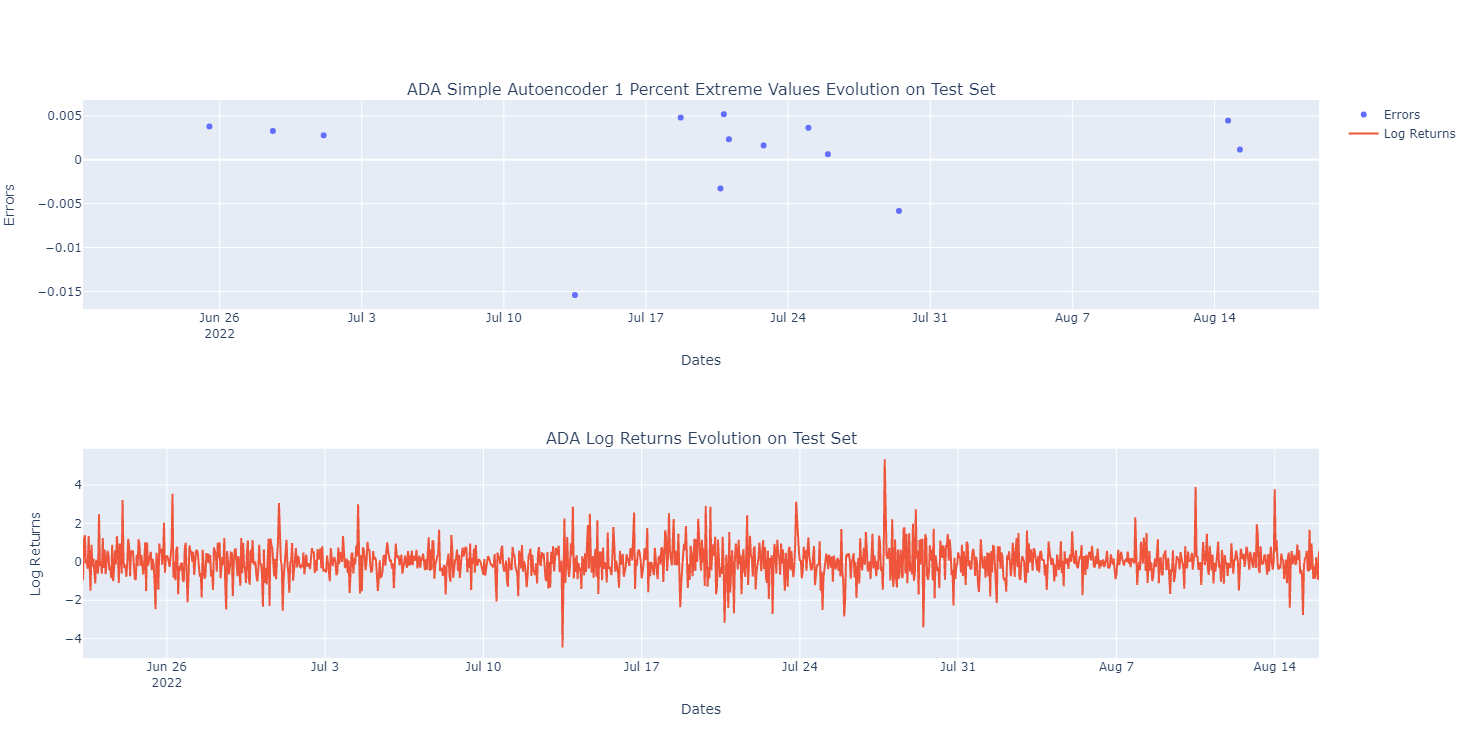}
\caption{ADA Simple Autoencoder 1 Percent Extreme Values Evolution on Test Set}
\label{fig:ADA Simple Autoencoder 1 Percent Extreme Values Evolution on Test Set}
\end{figure}

\begin{figure}[H]
    \centering
\includegraphics[width=0.8\columnwidth]{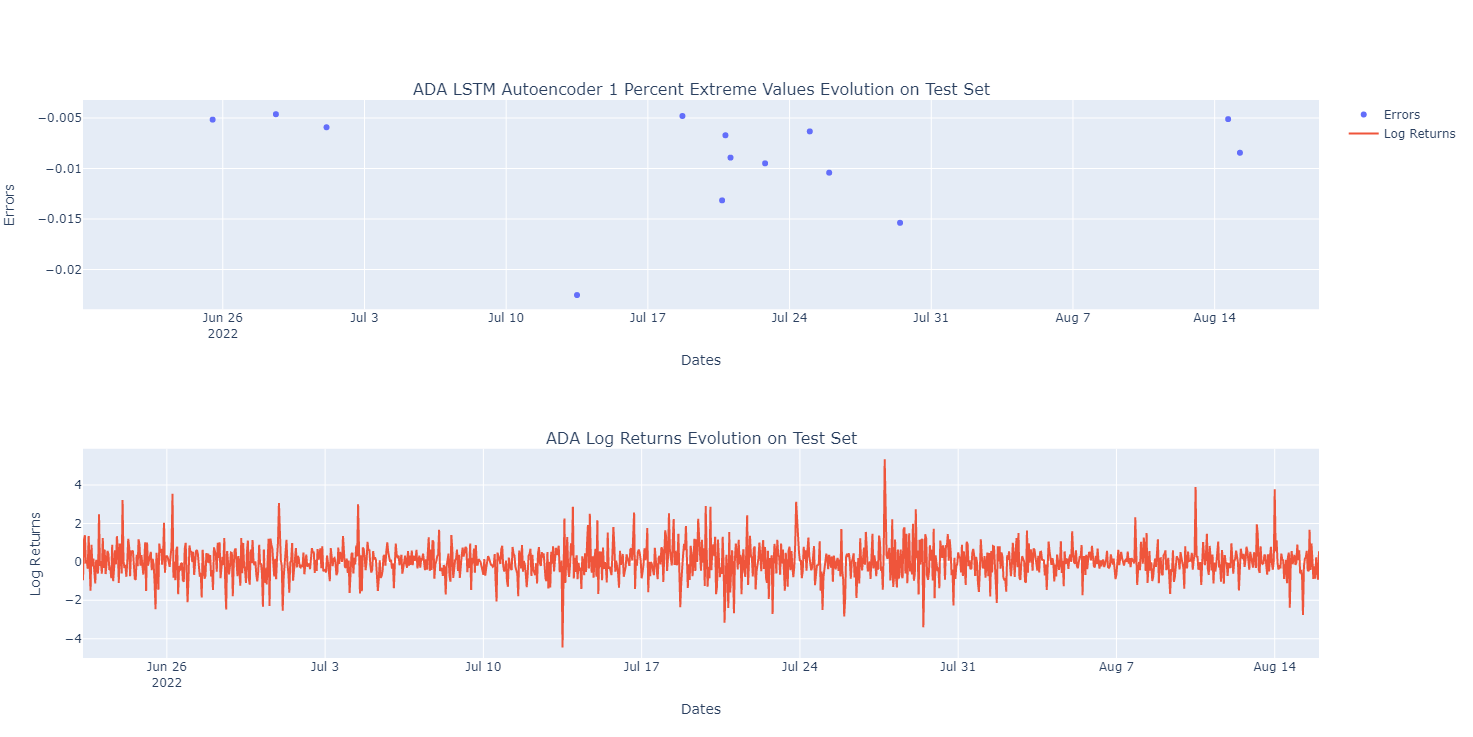}
\caption{ADA LSTM Autoencoder 1 Percent Extreme Values Evolution on Test Set}
\label{fig:ADA LSTM Autoencoder 1 Percent Extreme Values Evolution on Test Set}
\end{figure}

\begin{figure}[H]
    \centering
\includegraphics[width=0.8\columnwidth]{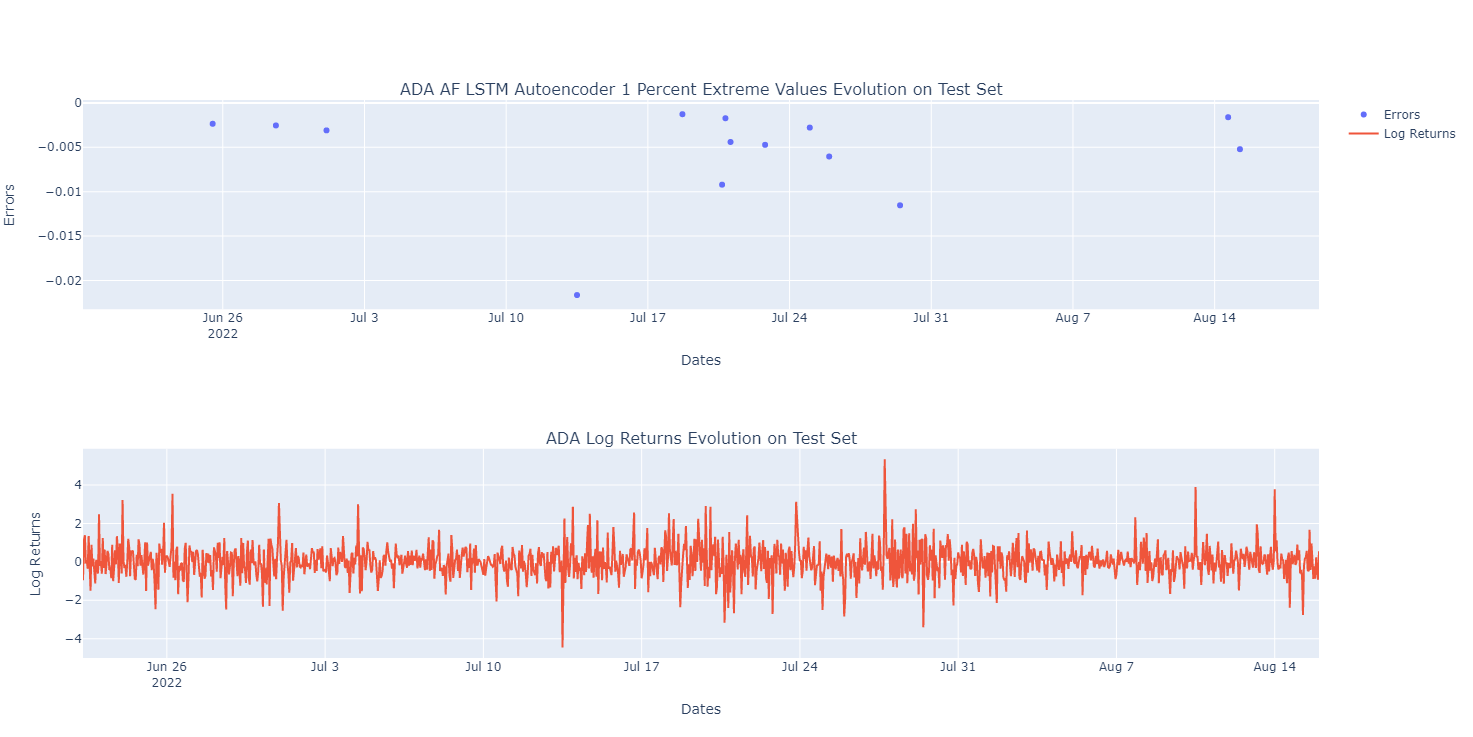}
\caption{ADA AF LSTM Autoencoder 1 Percent Extreme Values Evolution on Test Set}
\label{fig:ADA AF LSTM Autoencoder 1 Percent Extreme Values Evolution on Test Set}
\end{figure}

\end{document}